\documentclass[manuscript,screen]{acmart}
\renewcommand\footnotetextcopyrightpermission[1]{} 
\usepackage{colortbl}
\usepackage{graphicx}

\usepackage{hyperref}
\hypersetup{hypertex=true,
            colorlinks=true,
            linkcolor=red,
            anchorcolor=blue,
            citecolor=blue}
\usepackage{graphicx}%
\usepackage{multirow}%

\usepackage{amsmath,amsfonts}%
\usepackage{amssymb}
\usepackage{mathrsfs}%
\usepackage{xcolor}%
\usepackage{textcomp}%
\usepackage[normalem]{ulem}
\usepackage{microtype}

\usepackage{manyfoot}%
\usepackage{booktabs}%
\usepackage{algorithm}%
\usepackage{algorithmicx}%

\usepackage[noend]{algpseudocode}

\usepackage{listings}%
\usepackage{enumitem}
\usepackage{bbding}
\usepackage{multirow}
\usepackage{makecell}
\usepackage{soul}
\usepackage{tikz}

\definecolor{ceiling}{RGB}{210,42,33}
\definecolor{floor}{RGB}{40,159,7}
\definecolor{wall}{RGB}{162,215,224}
\definecolor{windows}{RGB}{113,158,201}
\definecolor{chair}{RGB}{202,205,76}
\definecolor{bed}{RGB}{224,190,159}
\definecolor{sofa}{RGB}{153,98,192}
\definecolor{table}{RGB}{23,124,181}
\definecolor{tvs}{RGB}{160,187,34}
\definecolor{furniture}{RGB}{225,126,18}
\definecolor{objects}{RGB}{199,174,221}

\definecolor{lightgreen}{rgb}{0.9,1,0.9}
\definecolor{lightred}{rgb}{1,0.9,0.9}

\definecolor{road}{RGB}{255,1,252}
\definecolor{sidewalk}{RGB}{76,0,74}
\definecolor{parking}{RGB}{255,149,255}
\definecolor{other_grnd}{RGB}{178,4,75}
\definecolor{building}{RGB}{255,198,0}
\definecolor{car}{RGB}{98,152,240}
\definecolor{truck}{RGB}{82,28,183}
\definecolor{bicycle}{RGB}{101,229,248}
\definecolor{motorcycle}{RGB}{35,57,148}
\definecolor{other_veh}{RGB}{100,82,245}
\definecolor{vegetation}{RGB}{2,176,0}
\definecolor{trunk}{RGB}{131,62,6}
\definecolor{terrain}{RGB}{153,238,85}
\definecolor{person}{RGB}{255,26,29}
\definecolor{bicyclist}{RGB}{255,40,202}
\definecolor{motorcycl}{RGB}{150,29,99}
\definecolor{fence}{RGB}{148,125,42}
\definecolor{pole}{RGB}{255,239,143}
\definecolor{traf_sign}{RGB}{248,2,3}

\newcommand{\thirdcolor}[1]{{#1}}

\AtBeginDocument{%
  }

\setcopyright{acmlicensed}
\copyrightyear{2018}
\acmYear{2018}
\acmDOI{XXXXXXX.XXXXXXX}

\acmConference[Conference acronym 'XX]{Make sure to enter the correct
  conference title from your rights confirmation emai}{June 03--05,
  2018}{Woodstock, NY}
\acmISBN{978-1-4503-XXXX-X/18/06}

\begin{document}

\title{Image Recognition  with Online Lightweight Vision Transformer: A Survey}


\author{Zherui Zhang}
\affiliation{%
  \institution{Beijing University of Posts and Telecommunications}
  \city{Beijing}
  \country{China}
}
\email{zzr787906410@bupt.edu.cn}

\author{Rongtao Xu}
\affiliation{%
  \institution{State Key Laboratory of Multimodal Artificial Intelligence Systems, Institute of Automation, Chinese Academy of Sciences}
  \city{Beijing}
  \country{China}
}
\email{xurongtao2019@ia.ac.cn}

\author{Jie Zhou}
\affiliation{%
  \institution{Beijing University of Posts and Telecommunications}
  \city{Beijing}
  \country{China}
}
\email{zhoujie8023@bupt.cn}

\author{Changwei Wang}
\affiliation{%
  \institution{Key Laboratory of Computing Power Network and Information Security, Ministry of Education, Shandong Computer Science Center, Qilu University of Technology, Shandong Provincial Key Laboratory of Computing Power Internet and Service Computing, Shandong Fundamental Research Center for Computer Science}
  \city{Jinan}
  \country{China}
}
\email{changweiwang@sdas.org}

\author{Xingtian Pei}
\affiliation{%
  \institution{Beijing University of Posts and Telecommunications}
    \city{Beijing}
  \country{China}
}
\email{2024111242@bupt.cn}

\author{Wenhao Xu}
\affiliation{%
  \institution{Beijing University of Posts and Telecommunications}
  \city{Beijing}
  \country{China}
}
\email{xuwenhao@bupt.edu.cn}

\author{Jiguang Zhang}
\affiliation{%
  \institution{State Key Laboratory of Multimodal Artificial Intelligence Systems, Institute of Automation, Chinese Academy of Sciences}
  \city{Beijing}
  \country{China}
}
\email{jiguang.zhang@ia.ac.cn}

\author{Li Guo}
\affiliation{%
  \institution{Beijing University of Posts and Telecommunications}
  \city{Beijing}
  \country{China}
}
\email{guoli@bupt.edu.cn}

\author{Longxiang Gao}
\affiliation{%
  \institution{Key Laboratory of Computing Power Network and Information Security, Ministry of Education, Shandong Computer Science Center, Qilu University of Technology, Shandong Provincial Key Laboratory of Computing Power Internet and Service Computing, Shandong Fundamental Research Center for Computer Science}
  \city{Jinan}
  \country{China}
}
\email{gaolx@sdas.org}

\author{Wenbo Xu}
\affiliation{%
  \institution{Beijing University of Posts and Telecommunications}
  \city{Beijing}
  \country{China}
}
\email{xuwb@bupt.edu.cn}

\author{Shibiao Xu}
\authornote{corresponding author}
\affiliation{%
  \institution{Beijing University of Posts and Telecommunications}
  \city{Beijing}
  \country{China}
}
\email{shibiaoxu@bupt.edu.cn}

\renewcommand{\shortauthors}{Zherui Zhang et al.}

\begin{abstract}
The Transformer architecture has achieved significant success in natural language processing, motivating its adaptation to computer vision tasks. 
Unlike convolutional neural networks, vision transformers inherently capture long-range dependencies and enable parallel processing, yet lack inductive biases and efficiency benefits, facing significant computational and memory challenges that limit its real-world applicability. 
This paper surveys various online strategies for generating lightweight vision transformers for image recognition, focusing on three key areas: \textbf{Efficient Component Design}, \textbf{Dynamic Network}, and \textbf{Knowledge Distillation}. 
We evaluate the relevant exploration for each topic on the ImageNet-1K benchmark, analyzing trade-offs among precision, parameters, throughput, and more to highlight their respective advantages, disadvantages, and flexibility. 
Finally, we propose future research directions and potential challenges in the lightweighting of vision transformers with the aim of inspiring further exploration and providing practical guidance for the community. 
Project Page: \url{https://github.com/ajxklo/Lightweight-VIT}

\end{abstract}



\maketitle

\section{\textbf{Introduction}}
\begin{figure}[!t]
    \centering
    \includegraphics[width=\linewidth]{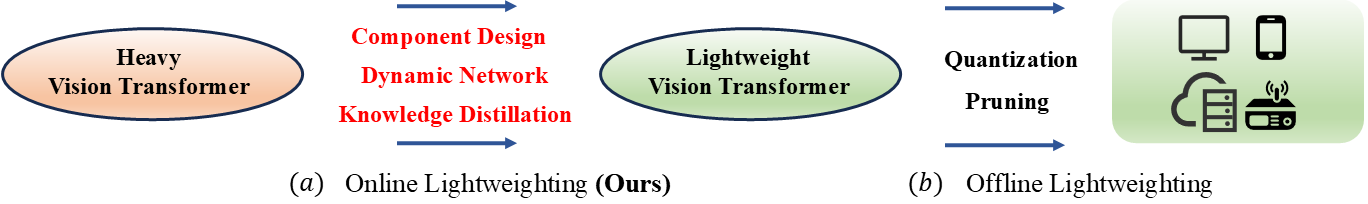}
    \caption{
    \textbf{ViT Lightweighting Overview.}
    We categorize existing techniques for alleviating the computational burden of ViT into two steps: online lightweighting and offline lightweighting. $(a)$ Online lightweighting techniques reduce the computational burden during training while ensuring the model structure or expressive capability is kept complete in a soft manner; $(b)$ Offline lightweighting techniques, such as pruning and quantization, trim unnecessary capacity from the model structure. In this paper, we focus more on online lightweighting techniques.
    }
    \label{fig:first_figure}
\end{figure}
The Vision Transformer~(ViT)~\citep{dosovitskiy2021imageworth16x16words,10.1145/3505244} rapidly emerge as powerful tools for image recognition, challenging the dominance of the traditional Convolutional Neural Network (CNN)~\citep{Xu2024HCFNetHC,xu2023wave,xu2023scd,Wang2023CNDescCN,Wang2023TreatingPG,Wang2025FocusOL,Xu2025RMLER,Wang2024ExploringID,Xu2024DomainFeatLL}. Using the self-attention mechanism of transformers in natural language processing, ViT models encode images as sequences of patches, allowing them to effectively capture global dependencies and contextual information~\citep{xu2024mrftrans}, yielding remarkable performance improvements in various vision tasks, including image generation~\citep{10.1145/3665869,10.1145/3586183.3606735,hatamizadeh2024diffitdiffusionvisiontransformers,dubey2024transformer,torbunov2023uvcgan,marnissi2023gan,gao2023masked,10.1145/3664647.3680768}, object detection~\citep{10.1145/3434398,10.1145/3519022,10.1145/3664598,amjoud2023object,zhang2023simple,kim2023region,gehrig2023recurrent,ye2023real,Zhao2024,Zhang2024-ly,Chen2023-cc}, Embodied Artificial Intelligence~\citep{ren2024infiniteworld,xu2025a0,liang2025structured, ma2025phyblock, han2025multimodal, zhang2025activevln, zhou2025mathcal, zhang2025robridge, xu20253d, chen2024constraint, yan2024instrugen, zhang2024navid} and semantic segmentation~\citep{thisanke2023semantic,Zhang2024-bn,ando2023rangevit,shi2023transformer,jain2023semask,Xu2023SpectralPT,xu2024-gba,Zhu2024,Zhang2024-ow,10.1145/3329784,Zhang2024MIMHDMS,Xu2024SkinFormerLS,Xu2024DefFusionDM}.
In addition, ViT models show remarkable precision in conventional image recognition benchmarks, achieving near-perfect validation results, such as surpassing an impressive precision $99\%$ on the CIFAR-10 dataset. 
Consequently, the limitations of small-scale datasets in fully demonstrating the performance superiority of ViT models lead to a shift towards larger benchmark datasets like ImageNet-1K~\citep{imagenet_cvpr09} for evaluating recognition tasks.

\begin{figure*}[tb]
    \centering
    \includegraphics[width=\textwidth]{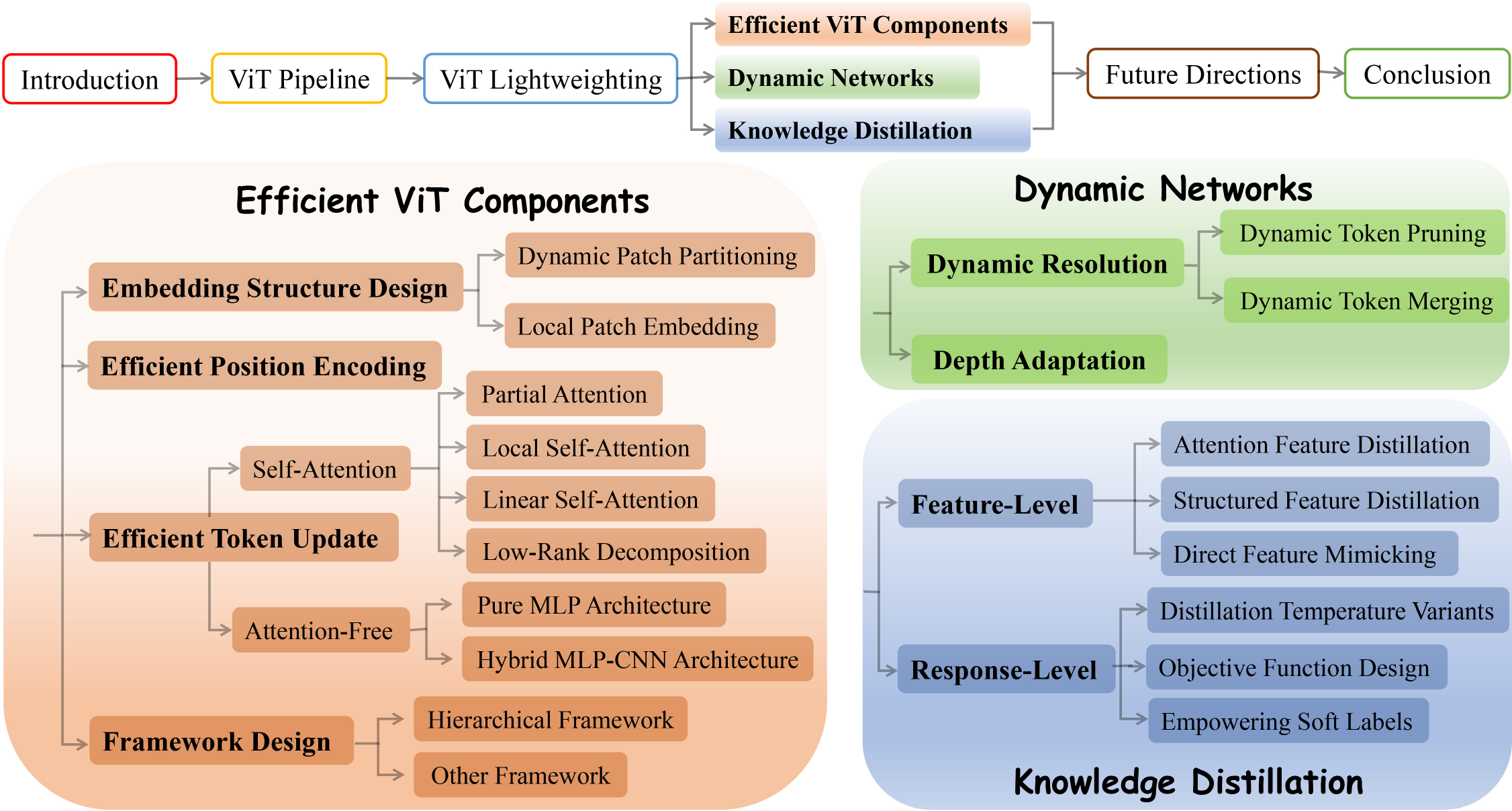}
    \caption{\textbf{Online ViT Lightweighting Overview.} 
    The general overview diagram of this paper shows the basic structure of the paper at the top, and at the bottom lists the methods collated in this paper regarding the online lightweighting of ViT models, mainly efficient ViT components, dynamic networks and knowledge distillation methods.
    }
    \label{fig:vis_token_OverFrame}
    \vspace{-0.4cm}
\end{figure*}

Despite ViT models impressive performance,~  they have significant drawbacks in model size and computational overhead. The increased parameters and complex operations slow training and inference and increase resource consumption, limiting practical deployment, especially in resource-constrained environments~\citep{tabani2021improving,chen2024comprehensivesurveymodelcompression}.
To address these issues, a variety of lightweighting techniques are proposed, aiming to reduce the computational burden without substantially compromising model performance.

This survey focuses on online lightweighting strategies\cite{10.1145/3657282,10.1145/3586074,10.1145/3530811} for ViT models, which differ from offline  post-processing methods such as pruning~\citep{yu2023x,liu2024evolutionvit,yu2023unified} and quantization~\citep{yu2023boost,li2023repq,du2024modelquantizationhardwareacceleration}. Online techniques aim to optimize the model during training, either by improving the design of internal components~\citep{han2023flatten,liu2023efficientvit}, transferring knowledge~\citep{zhao2023cumulative,cai2023task,Zhang2025CAEDFKDBT,10980222}, or dynamically adjusting the computational path of the network based on the complexity of the input~\citep{rao2023dynamic,xia2023dat++}. The goal is to achieve a ``\textbf{soft}" lightweighting, maintaining the model architecture and expressive capacity while enhancing efficiency. As illustrated in Figure~\ref{fig:first_figure}, online lightweighting techniques are seen as a precursor to more aggressive offline post-processing methods.

We categorize online lightweighting techniques for ViT models into three main topics: efficient component design, dynamic network, and knowledge distillation~(KD).
These topics exploit the ViT property on token-based input to achieve unique efficiency gains. 
For example, token input updates can be handled using native self-attention, refined linear attention, or even attention-free~\citep{lai2023mask,buoy2023vitstr} mechanisms. 
To address spatial redundancy~\citep{chen2022principle,zhang2022pan}, dynamic token pruning~\citep{pan2021ia,rao2023dynamic} and token merging~\citep{feng2023efficient,kim2024token} strategies are used, offering a more granular approach compared to traditional downsampling of CNN models. 
Integrating these techniques allows for customized,  lightweight strategies developed for real-world applications.
Our contributions in this survey are as follows:
\begin{itemize}
    \item \textbf{Comprehensive Overview}. We provide a broad survey  of online lightweighting techniques for vision transformer, distinguishing them from offline post-processing techniques, and categorize them into three key topics: efficient component design, dynamic network, and knowledge distillation, with each topic discussed in the context of input flow forwarding.
    \item \textbf{Trade-Off Presentation.} Using ImageNet-1K~\citep{imagenet_cvpr09} as the benchmark, we present a detailed analysis of the trade-offs between precision and efficiency for various online lightweighting strategies, offering both qualitative and quantitative insights.
    \item  \textbf{Future Directions}. 
    We identify the limitations of current online lightweighting techniques, highlight common assumptions that may need reconsideration, and propose promising research directions, including privacy-sensitive models.
\end{itemize}

\begin{table*}[!ht]
\centering
\renewcommand{\arraystretch}{1.33} 
\caption{Some Simplified Forms of symbols in this paper.}
\resizebox{\textwidth}{!}{
\begin{tabular}{c|c|c|c|c}
\toprule[1.5px]
\multicolumn{5}{c}{\textbf{Usage}} \\
\hline
\rowcolor{gray!20} 
\textbf{CNN} & \textbf{KD} & \textbf{ViT} & \textbf{Patch or Token} & \textbf{MLP} \\
\hline
Convolutional Neural Network & Knowledge Distillation & Vision Transformer & ViT-specific Input Form & Multi-Layer Perceptron \\
\hline
\rowcolor{gray!20}
\textbf{FFN} & \textbf{SA} & \textbf{MHSA} & \textbf{PE} & \textbf{LN} \\
\hline
Feed-Forward Network & Self-Attention & Multi-Head Self-Attention & Positional Encoding & Layer Normalization \\
\hline
\rowcolor{gray!20}
\textbf{BN} & \textbf{CE} & \textbf{KL} & \textbf{SOTA} & \textbf{FLOPs} \\
\hline
Batch Normalization & Cross Entropy & Kullback-Leibler & State Of The Art & Floating Point Operations per Second \\
\hline
\hline
\multicolumn{5}{c}{\textbf{Symbol}} \\
\hline
\rowcolor{gray!20}
\textbf{Q} & \textbf{V} & \textbf{K} & \textbf{C} & \textbf{$\Omega$} \\
\hline
Query Matrix & Value Matrix & Key Matrix & Embedding Dimension & Computation Complexity \\
\bottomrule[1.5px]
\end{tabular}
}
\label{tab:my_label}
\end{table*}

\noindent 
{{\textbf{Difference from Other Surveys}}} This survey focuses mainly on online lightweighting techniques of ViT models and provides comprehensive experimental reports to uniformly examine the trade-offs existing strategies make between precision and efficiency.

\noindent 
{{\textbf{Survey Overview}}} 
This survey uses simplified notation from Table~\ref{tab:my_label}. Figure~\ref{fig:vis_token_OverFrame} shows the structure: Section~\ref{Original ViT Pipeline} introduces the vision transformer concept and pipeline. We then present online lightweight strategies  on three main topics. 
The first topic, discussed in Section~\ref{Efficient ViT Components}, explores the \textbf{efficient component} design within ViT. 
The second topic, presented in Section~\ref{Dynamic Networks}, explores \textbf{dynamic networks} based on input complexity, including dynamic resolution and depth adaptation. 
The third topic, in Section~\ref{Knowledge Distillation}, focuses on the use of the unique advantages of ViT for \textbf{knowledge distillation}.
For each topic, we report from a unified perspective on the results of image recognition performance evaluations on the ImageNet~\citep{imagenet_cvpr09} benchmark for the methods involved, analyzing their advantages and disadvantages, as well as flexibility.
Finally, in Section~\ref{Future Research Directions}, we offer a discussion on promising directions for future exploration.

\section{\textbf{Original ViT Pipeline}}
\label{Original ViT Pipeline}

\begin{figure*}[!t]
    \centering
    \includegraphics[width=\textwidth]{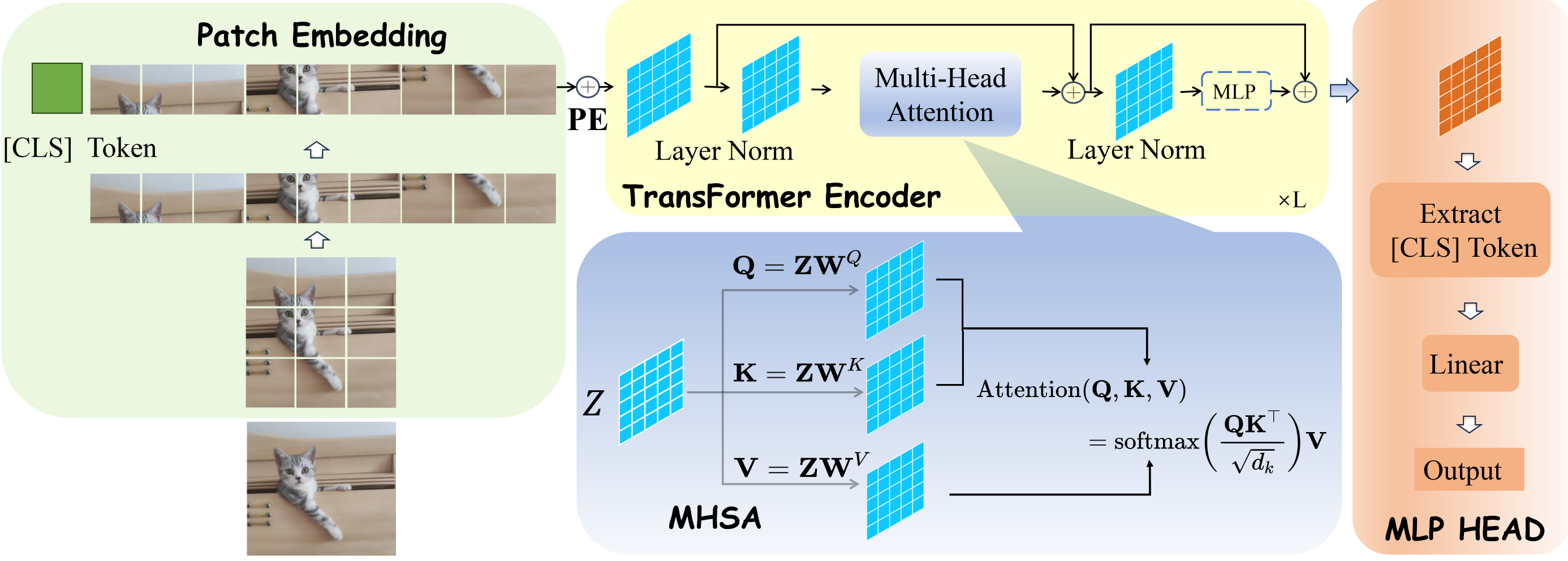}
    \caption{\textbf{Original ViT Overview.}
    The pipeline of the original ViT for image recognition is divided into four main components. Firstly, Patch Embedding is used to obtain input in token form. Secondly, there is a stack of Transformer layers, which includes the Multi-Head Self-Attention (MHSA) mechanism, facilitating ViT global comprehension ability. Finally, a Multi-Layer Perceptron (MLP) structure is employed to project into category space.
    }
    \label{fig:vis_token_VIT2}
    \vspace{-0.4cm}
\end{figure*}

As shown in Figure~\ref{fig:vis_token_VIT2}, given an input image $I$ with resolution $H \times W$, the first step in the ViT pipeline involves dividing the image into non-overlapping $N = (H/P) \times (W/P)$ patches, each of size $P \times P$, where $P$ represents the patch size. Each patch $\mathbf{x}_p \in \mathbb{R}^{P \times P \times C}$ undergoes a flattening operation to obtain a linear embedding $\mathbf{x}_p \in \mathbb{R}^{P^2C}$, followed by a linear projection to obtain the patch embedding $\mathbf{z}_p^0 = \mathbf{E}\mathbf{x}_p$, where $\mathbf{E} \in \mathbb{R}^{D \times P^2C}$ is a learnable projection matrix and $D$ is the embedding dimension.

To incorporate positional information, a positional embedding $\mathbf{E}_{pos} \in \mathbb{R}^{(N+1) \times D}$ is added to the patch embeddings. The resulting sequence of embedded patches $\mathbf{Z}_0 = [\mathbf{z}_{cls}; \mathbf{z}_1^0; \ldots; \mathbf{z}_N^0] + \mathbf{E}_{pos}$ serves as the input to the transformer encoder, where $\mathbf{z}_{cls} \in \mathbb{R}^D$ is a learnable [CLS] token that facilitates recognition.

The transformer encoder consists of $L$ layers, each comprising a Multi-Head Self-Attention (MHSA) block, a Feed-Forward Network (FFN) or Multilayer Perceptron (MLP). 
In the MHSA block, the input sequence $\mathbf{Z}_{l-1}$ is linearly projected to obtain the query $\mathbf{Q}$, key $\mathbf{K}$ and value $\mathbf{V}$ using the learned projection matrices $\mathbf{W}_q, \mathbf{W}_k, \mathbf{W}_v \in \mathbb{R}^{D \times D_h}$, respectively. The attention weights are computed as $\text{Attention}(\mathbf{Q}, \mathbf{K}, \mathbf{V}) = \text{softmax}(\frac{\mathbf{Q}\mathbf{K}^T}{\sqrt{D_h}})\mathbf{V}$, where $D_h$ is the dimension of each head. The outputs of the $H$ attention heads are concatenated and linearly projected using $\mathbf{W}_o \in \mathbb{R}^{HD_h \times D}$ to obtain the MSA output.

The FFN consists of two linear transformations with an activation function between, applied positionally to the MHSA output: $\text{FFN}(\mathbf{Z}) = \text{ReLU}(\mathbf{Z}\mathbf{W}_1 + \mathbf{b}_1)\mathbf{W}_2 + \mathbf{b}_2$, \ where $\mathbf{W}_1 \in \mathbb{R}^{D \times D_{FFN}}$, $\mathbf{b}_1 \in \mathbb{R}^{D_{FFN}}$, $\mathbf{W}_2 \in \mathbb{R}^{D_{FFN} \times D}$, and $\mathbf{b}_2 \in \mathbb{R}^D$ are learnable parameters.
Each transformer encoder layer incorporates residual connections and layer normalization~(LN), resulting in the output:
$ \mathbf{Z}_l  =  \text{LN} 
        (\mathbf{Z}_{l-1} + \text{FFN}(\text{LN}(\mathbf{Z}_{l-1} + \text{MSA}(\mathbf{Z}_{l-1})))).$
        
After passing through the $L$ transformer encoder layers, the final output sequence $\mathbf{Z}_L = [\mathbf{z}_{cls}^L; \mathbf{z}_1^L; \ldots; \mathbf{z}_N^L]$ is obtained. The output of the [CLS] token $\mathbf{z}_{cls}^L$ is used for recognition by applying a linear transformation followed by softmax activation: $\mathbf{y} = \text{softmax}(\mathbf{W}_{cls}\mathbf{z}_{cls}^L + \mathbf{b}_{cls})$, where $\mathbf{W}_{cls} \in \mathbb{R}^{D \times K}$, $\mathbf{b}_{cls} \in \mathbb{R}^K$, and $K$ is the number of classes.

\section{\textbf{Efficient ViT Components}}
\label{Efficient ViT Components}
As the desire for improved image recognition precision continues to grow, challenges arise from image quality, object characteristics, and potential dependencies. The vision transformer stands out as particularly well suited for such tasks because of the unique representation advantages and robust adaptability to global understanding. 
Consequently, novel designs for various components of ViT emerge in abundance, accompanied by additional training and inference costs.

In this section, we present the efforts towards online lightweight design of each component in the order in which it processes the input image within the network. 
This includes \textbf{Embedding Structure Design} in Section~\ref{Embedding Structure Design}, \textbf{Efficient Position Encoding} in Section~\ref{Efficient Position Encoding}, \textbf{Efficient Token Update} in Section~\ref{Efficient Token Update}, and \textbf{Framework Design} in Section~\ref{Framework Design}.

\subsection{\textbf{Embedding Structure Design}}
\label{Embedding Structure Design}
Patch embedding serves as the initial stage in the ViT pipeline, responsible for converting the input image into a sequence of tokens that can be processed by the transformer encoder. This stage plays a crucial role in determining the efficiency of the overall model, as the number of patches directly influences the computational complexity of subsequent self-attention layers.
Specifically, we discuss two prominent directions: \textbf{Dynamic Patch Partitioning~(DPP)}  and \textbf{Local Patch Embedding~(LPE)}. 
DPP explores adaptive partitioning of the input image to generate effective patches at multiple scales and multiple views. 
On the other hand, LPE leverages CNN models to extract rich local information.

\begin{table*}[t]
\caption{
A quantitative comparative analysis is conducted on the classic methods involved in \textbf{Embedding Structure Design}~(Section~\ref{Embedding Structure Design}), using ImageNet-1K~\citep{imagenet_cvpr09} as the benchmark, with the resolution standardized to $224 \times 224$.
\textsuperscript{†} denotes that the CrossViT model is modified by replacing the linear patch embedding in ViT with three convolutional layers as patch labelers.
The best-performing method is highlighted in bold, and the runner-up is underlined.
}

    \begin{minipage}[t]{0.49\linewidth}
    \renewcommand{\arraystretch}{1.33} 
    \centering
    \resizebox{\linewidth}{!}{
        \begin{tabular}{lcrrr}
        \bottomrule[1.5px]
         \textbf{Model} &\textbf{Year} & \textbf{\makecell[r]{Params \\ (M)$\downarrow$}}  & \textbf{\makecell[r]{FLOPs \\ (G)$\downarrow$}}  & \textbf{\makecell[r]{Top-1 Acc. \\ (\%)$\uparrow$}}\\
    \hline
    \tikz\fill[sofa] (0,0) -- (8pt,0) -- (4pt,8pt) -- cycle;\:CrossViT-Ti$|$DeiT-Ti~\cite{chen2021crossvit} &2021& 6.9 & 1.6 & 73.4\\
    \tikz\fill[sofa] (0,0) -- (8pt,0) -- (4pt,8pt) -- cycle;\:CrossViT-9$|$DeiT-Ti~\cite{chen2021crossvit} &2021& 8.6 & 1.8 & 73.9\\
    \tikz\fill[sofa] (0,0) -- (8pt,0) -- (4pt,8pt) -- cycle;\:CrossViT-9\textsuperscript{†}$|$DeiT-Ti~\cite{chen2021crossvit} &2021& 8.8 & 2.0 & 77.1\\
    \tikz\fill[sofa] (0,0) -- (8pt,0) -- (4pt,8pt) -- cycle;\:CrossViT-S$|$DeiT-S~\cite{chen2021crossvit} &2021& 26.7 & 5.6 & 81.0\\
    \tikz\fill[sofa] (0,0) -- (8pt,0) -- (4pt,8pt) -- cycle;\:CrossViT-15$|$DeiT-S~\cite{chen2021crossvit} &2021& 27.4 & 5.8 & 81.5\\
    \tikz\fill[sofa] (0,0) -- (8pt,0) -- (4pt,8pt) -- cycle;\:CrossViT-15\textsuperscript{†}$|$DeiT-S~\cite{chen2021crossvit} &2021& 28.2 & 6.1 & 82.3\\
    \tikz\fill[sofa] (0,0) -- (8pt,0) -- (4pt,8pt) -- cycle;\:CrossViT-18$|$DeiT-B~\cite{chen2021crossvit} &2021& 43.3 & 9.0 & 82.5\\
    \tikz\fill[sofa] (0,0) -- (8pt,0) -- (4pt,8pt) -- cycle;\:CrossViT-18\textsuperscript{†}$|$DeiT-B~\cite{chen2021crossvit} &2021& 44.3 & 9.5 & 82.8\\
    \tikz\fill[sofa] (0,0) -- (8pt,0) -- (4pt,8pt) -- cycle;\:CrossViT-B$|$DeiT-B~\cite{chen2021crossvit} &2021& 104.7 & 21.2 & 82.2\\
    \hline
    \rowcolor{gray!20} \tikz\fill[truck] (0,0) -- (8pt,0) -- (4pt,8pt) -- cycle;\:IA-RED$^2|$DeiT-S~\cite{pan2021ia} &2021& \thirdcolor{22.1} & - & 79.1\\
    \rowcolor{gray!20} \tikz\fill[truck] (0,0)  -- (8pt,0) -- (4pt,8pt) -- cycle;\:IA-RED$^2|$DeiT-B~\cite{pan2021ia} &2021& \thirdcolor{86.6} & 11.8 & 80.3\\
    \hline
    \tikz\fill[trunk] (0,0) -- (8pt,0) -- (4pt,8pt) -- cycle;\:CvT-13~\cite{Wu_2021_ICCV} & 2021 & 20.0 & 4.5 & 81.6\\
    \tikz\fill[trunk] (0,0) -- (8pt,0) -- (4pt,8pt) -- cycle;\:CvT-21~\cite{Wu_2021_ICCV} & 2021 & 32.0 & 7.1 & 82.5\\
     \hline
    \rowcolor{gray!20}  \tikz\fill[gray] (0,0) -- (8pt,0) -- (4pt,8pt) -- cycle;\:MobileViT-XS~\cite{mehta2022mobilevitlightweightgeneralpurposemobilefriendly} & 2022 & 2.3 & 0.7 & 74.8\\
    \rowcolor{gray!20}  \tikz\fill[gray] (0,0) -- (8pt,0) -- (4pt,8pt) -- cycle;\:MobileViT-S~\cite{mehta2022mobilevitlightweightgeneralpurposemobilefriendly} & 2022 & 5.6 & - & 78.4\\
    \hline
    \tikz\fill[objects] (0,0) circle (3.5pt);\:Mobile-Former-26M~\cite{chen2022mobileformer} & 2022 & 3.2 & - & 60.4\\
    \tikz\fill[objects] (0,0) circle (3.5pt);\:Mobile-Former-52M~\cite{chen2022mobileformer} & 2022 & 3.5 & - & 68.7\\
    
    \bottomrule[1.5px]
    \end{tabular}
        }
    \end{minipage}
    \hfill
     \begin{minipage}[[t]{0.49\linewidth}
     \renewcommand{\arraystretch}{1.32} 
     \centering
     \resizebox{\linewidth}{!}{
        \begin{tabular}{lcrrr}
        \bottomrule[1.5px]
         \textbf{Model} &\textbf{Year} & \textbf{\makecell[r]{Params \\ (M)$\downarrow$}}  & \textbf{\makecell[r]{FLOPs \\ (G)$\downarrow$}}  & \textbf{\makecell[r]{Top-1 Acc. \\ (\%)$\uparrow$}}\\
    \hline
    \tikz\fill[objects] (0,0) circle (3.5pt);\:Mobile-Former-96M~\cite{chen2022mobileformer} & 2022 & 4.6 & - & 72.8\\
    \tikz\fill[objects] (0,0) circle (3.5pt);\:Mobile-Former-151M~\cite{chen2022mobileformer} & 2022 & 7.6 & - & 75.2\\
    \tikz\fill[objects] (0,0) circle (3.5pt);\:Mobile-Former-214M~\cite{chen2022mobileformer} & 2022 & 9.4 & - & 76.7\\
    \tikz\fill[objects] (0,0) circle (3.5pt);\:Mobile-Former-294M~\cite{chen2022mobileformer} & 2022 & 11.4 & - & 77.9\\
    \tikz\fill[objects] (0,0) circle (3.5pt);\:Mobile-Former-508M~\cite{chen2022mobileformer} & 2022 & 14.0 & - & 79.3\\
    \hline
    \rowcolor{gray!20} \tikz\fill[ceiling] (0,0) circle (3.5pt);\:SSA-T~\cite{ren2022shunted} & 2022 & 11.3 & 2.1 & 79.8\\
    \rowcolor{gray!20} \tikz\fill[ceiling] (0,0) circle (3.5pt);\:SSA-S~\cite{ren2022shunted} & 2022 & 22.4 & 4.9 & 82.9\\
    \rowcolor{gray!20} \tikz\fill[ceiling] (0,0) circle (3.5pt);\:SSA-B~\cite{ren2022shunted} & 2022 & 39.6 & 8.1 & \underline{84.0}\\
    \rowcolor{gray!20} \tikz\fill[ceiling] (0,0) circle (3.5pt);\:SSA-L~\cite{ren2022shunted} & 2022 & 81.2 & 14.9 & \textbf{84.6}\\
   \hline
   \tikz\fill[trunk] (0,0) rectangle (8pt,6pt);\:DPS-ViT$|$DeiT-Ti~\cite{tang2022patch} &2022& \thirdcolor{5.7} & 0.6 & 72.1\\
    \tikz\fill[trunk] (0,0) rectangle (8pt,6pt);\:DPS-ViT$|$T2T-ViT~\cite{tang2022patch} &2022& \thirdcolor{21.5} & 3.1 & 81.3\\
   \tikz\fill[trunk] (0,0) rectangle (8pt,6pt);\:DPS-ViT$|$DeiT-S~\cite{tang2022patch} &2022& \thirdcolor{22.1} & 2.6 & 79.4\\
   \tikz\fill[trunk] (0,0) rectangle (8pt,6pt);\:DPS-ViT$|$DeiT-B~\cite{tang2022patch} &2022& \thirdcolor{86.6} & 9.4 & 81.6\\
    \hline
    \rowcolor{gray!20} \tikz\fill[bed] (0,0) rectangle  (6.5pt,6.5pt);RepViT-M0.9~\cite{Wang_2024_CVPR} &2024 &5.4 &1.6 &78.7\\
    \rowcolor{gray!20} \tikz\fill[bed] (0,0) rectangle  (6.5pt,6.5pt);RepViT-M1.0~\cite{Wang_2024_CVPR} &2024 &6.8 &2.2 &80.0\\
    \rowcolor{gray!20} \tikz\fill[bed] (0,0) rectangle  (6.5pt,6.5pt);RepViT-M1.1~\cite{Wang_2024_CVPR} &2024 &8.2 &2.6 &80.7\\
    \rowcolor{gray!20} \tikz\fill[bed] (0,0) rectangle  (6.5pt,6.5pt);RepViT-M2.3~\cite{Wang_2024_CVPR} &2024 &22.9 &9 &83.3\\
    \bottomrule[1.5px]
        \end{tabular}
        }
    \end{minipage}
    \label{table:patch embedding}
\end{table*}

\subsubsection{\textbf{Dynamic Patch Partitioning}}
DPP adapts to the computational demands by employing patches of varying sizes or dynamically adjusting patch dimensions. CrossViT\footnote{ \url{https://github.com/IBM/CrossViT}}~\citep{chen2021crossvit} employs a dual-branch Transformer with multi-scale patches, while SSA~\citep{ren2022shunted} merges tokens to represent larger objects efficiently. TopFormer\footnote{ \url{https://github.com/hustvl/TopFormer}}~\citep{zhang2022topformer} integrates multi-scale features using token and semantic pyramid modules, and T2T-ViT\footnote{ \url{https://github.com/yitu-opensource/T2T-ViT}}~\citep{Yuan_2021_ICCV} adopts a hierarchical approach to model local structures. Dynamic patch partitioning techniques, such as IA-RED$^2$~\citep{pan2021ia}, DynamicViT~\citep{rao2021dynamicvit}, and~\citep{tang2022patch}, eliminate redundant or non-essential patches to optimize computational efficiency.

\subsubsection{\textbf{Local Patch Embedding}}
LPE incorporates convolutional layers into the patch embedding process, leveraging its local feature extraction capabilities while preserving the ViT  global context understanding. 
MobileViT~\citep{mehta2022mobilevitlightweightgeneralpurposemobilefriendly} uses convolutional layers to encode pixel information within non-overlapping patches. 
MobileFormer~\citep{chen2022mobileformer} integrates a lightweight MobileNet, using global tokens to fuse local and global features. 
CvT~\citep{Wu_2021_ICCV} employs a hierarchical design with convolutional token embedding layers at each stage.

\subsubsection{\textbf{Summary: Embedding Structure Design  }}
The traditional global patch encoding strategy does not adequately capture local information from images, affecting the precision of fine-grained image recognition on ImageNet~\citep{imagenet_cvpr09}, while a dense patch encoding approach incurs a higher computational burden. Two directions for improvement under this topic involve multi-scale and multi-perspective adaptive partitioning of input images, combined with CNNs to extract rich local features, enhancing the quality and efficiency of patch embedding. In Table ~\ref{table:patch embedding}, we report the relevant experimental results, where the following phenomena can be observed: 
\textit{\textbf{(i).}} The CrossViT~\citep{chen2021crossvit} series introduces dynamic patch partitioning and local partition embedding mechanisms, significantly outperforming the traditional DeiT model, with competitive FLOPs and an accuracy improvement from $81.0\% \rightarrow 82.3\%$;
\textit{\textbf{(ii).}} Lightweight variants such as  
MobileViT~\citep{mehta2022mobilevitlightweightgeneralpurposemobilefriendly} and CvT~\citep{Wu_2021_ICCV} achieve an excellent balance between accuracy and efficiency, for instance, MobileViT-S~\citep{mehta2022mobilevitlightweightgeneralpurposemobilefriendly} achieves an accuracy of $78.4\%$ with \ just $5.6$M parameters; 
\textit{\textbf{(iii).}}  State-of-the-art~(SOTA)  methods like SSA~\citep{ren2022shunted}  and DPS-ViT~\citep{tang2022patch} achieve a win-win in both accuracy and efficiency, with SSA-B~\citep{ren2022shunted} achieving a high accuracy of $84.0\%$ with only $39.6$M parameters.

\begin{figure*}[tb]
    \centering
    \includegraphics[width=\linewidth]{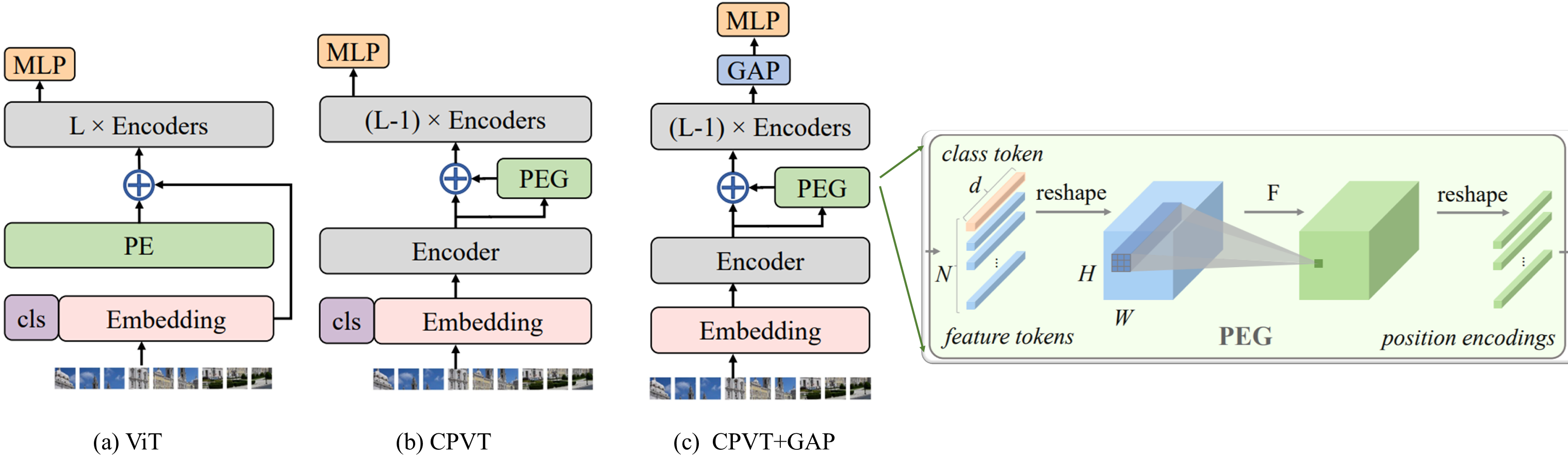}
    \caption{
    \textbf{How Positional Encoding Works.}
   A comparative illustration~\citep{chu2023conditional}  of the  vision transformer and its modified counterpart is presented. $(a)$ depicts the ViT model that employs explicit learnable positional encodings. In contrast, $(b)$ showcases the Conditional Positional Vision Transformer (CPVT) utilizing conditional positional encoding generated by the proposed Position Encoding Generator (PEG) plugin, which serves as the default configuration. $(c)$ illustrates the CPVT-GAP variant, which, in the absence of a [CLS] token, performs global average pooling (GAP) across all sequence elements. 
    }
    \label{fig:Positional Encoding}
\end{figure*}

\subsection{\textbf{Efficient Position Encoding}}
\label{Efficient Position Encoding}
After converting the image input into a sequence of patches, it is necessary to encode the spatial information lost during patch partitioning. 
This is where Position Encoding~(PE) comes into play, as shown in Figure~\ref{fig:Positional Encoding}~$(a)$.
Traditional absolute positional encodings, such as the widely used sinusoidal method:
\begin{equation}
    \begin{aligned}
        \text{PE}=\left\{\begin{array}{lll}
\sin \left(\frac{\text { pos }}{10000^{2 k / d}}\right) & \text { if } & i=2 k \\
\cos \left(\frac{\text { pos }}{10000^{2 k / d}}\right) & \text { if } & i=2 k+1
\end{array}\right.
    \end{aligned}
\end{equation}
where \(k=1, \ldots, d / 2\), \(i\) and \(d\) denote the index and length of the vector respectively, and ``pos” denotes the position of the element in the sequence.
While simple, this method suffers from limitations such as lack of adaptability to variable input sizes or non-available prior information. 
Recent research has shifts towards more efficient and flexible position encoding techniques that can effectively capture spatial relationships while minimizing computational overhead.

One promising direction is to incorporate relative positional information directly into the attention mechanism. RoPE \citep{heo2024rotary} achieves this by integrating query and key vectors, improving the ability of self-attention to encode spatial relationships. Similarly, PoSGU \citep{Wang_2022} embeds positional information within the token mixing layer, maintaining expressiveness while reducing parameter overhead. Another approach leverages efficient computation methods for relative positional embeddings. LISA \citep{kwon2022lightweight} employs the Fast Fourier Transform (FFT) to approximate relative positional embeddings, achieving log-linear complexity. FIRE \citep{li2024functional} introduces gradient interpolation techniques for relative encoding, and LRPE \citep{qin2023relative} proposes efficient linear complexity encodings.

Furthermore, adapting positional encoding to varying input contexts shows significant promise. For example, as shown in Figure~\ref{fig:Positional Encoding}, CPVT~\citep{chu2023conditional} dynamically incorporate positional information by conditioning image patch content, thus enhancing model flexibility and performance. This is particularly beneficial for tasks such as extrapolating lengths in transformers, significantly improving medical segmentation tasks with Polyp-ViT \citep{Sikkandar2024UtilizingAD}.

\subsubsection{\textbf{Summary: Efficient Position Encoding}}
ViT is highly dependent on the self-attention mechanism for outstanding global understanding, which inherently lacks the ability to perceive the sequential or spatial relationships of the input data.
Effectively encoding positional information can enhance performance in several aspects: 
\textbf{\textit{(i).}} Facilitates the ViT model in comprehending the spatial structure of images, enabling it to differentiate objects and their positions within the frame, which is particularly crucial for ImageNet~\citep{imagenet_cvpr09}, which contains a vast number of objects with diverse spatial configurations; 
\textit{\textbf{(ii).}}   Allowing the ViT model to focus more effectively on relevant regions of the image;
\textit{\textbf{(iii).}}   Eliminates ambiguity in feature representations by ensuring that the position of each pixel or patch is known; 
\textit{\textbf{(iv).} }   Accelerates the ViT model convergence speed during the training process by providing clear spatial signals, making the learning of large-scale image recognition datasets more reliable.

\subsection{\textbf{Efficient Token Update}}
\label{Efficient Token Update}
\subsubsection{\textbf{Self-Attention}}
\label{Self-Attention}
With the spatial context re-introduced, the core component of the ViT architecture comes into play: Self-Attention (SA). 
This mechanism empowers the model to capture long-range dependencies among patche inputs, enabling the comprehensive understanding of the image. 
However, the computational complexity of standard self-attention grows quadratically with the number of tokens, posing a significant bottleneck for resource-constrained applications.

To mitigate this issue, researchers have made significant efforts to develop efficient self-attention mechanisms that maintain the ability of the model to capture global dependencies while reducing computational cost. 
This section explores four prominent directions in efficient self-attention design: \textbf{Partial Attention}, \textbf{Local Self-Attention}, \textbf{Linear Self-Attention}, and \textbf{Low-rank Decomposition}.

\begin{figure*}
    \centering
    \includegraphics[width=1.0\linewidth]{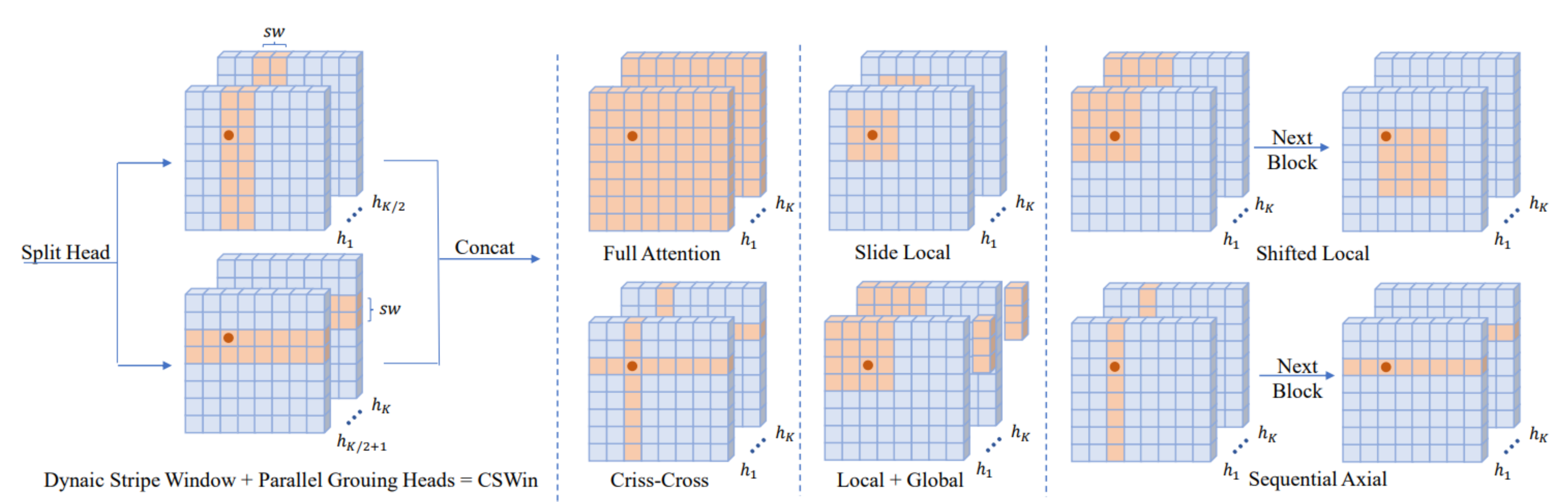}
    \caption{
    \textbf{Improvements in Self-Attention.}
    CSwin~\citep{dong2022cswin} splits multi-heads ({$h_1$, \dots , $h_K$}) into two groups and performs self-attention in horizontal and vertical stripes simultaneously. Then, CSwin adjusts the stripe width according to the network depth, which achieves a better trade-off between computational cost and capability.}
    \label{fig:cwwin}
\end{figure*}

\begin{table*}[h]
\caption{
A quantitative comparative analysis is conducted on the classic methods involved in \textbf{Self-Attention Optimization}~(Section~\ref{Self-Attention}), using ImageNet-1K~\citep{imagenet_cvpr09} as the benchmark, with the resolution standardized to $224 \times 224$.
The best-performing method is highlighted in bold, and the runner-up is underlined.
}
    
    \begin{minipage}[t]{0.49\linewidth}
    \renewcommand{\arraystretch}{1.33} 
    \centering
    \resizebox{\linewidth}{!}{
        \begin{tabular}{lcrrr}
        \bottomrule[1.5px]
         \textbf{Model} &\textbf{Year} & \textbf{\makecell[r]{Params \\ (M)$\downarrow$}}  & \textbf{\makecell[r]{FLOPs \\ (G)$\downarrow$}}  & \textbf{\makecell[r]{Top-1 Acc. \\ (\%)$\uparrow$}}\\
    \hline
    \tikz\fill[ceiling] (0,0) circle (3.5pt);\:GFNet-Ti~\cite{rao2021global} & 2021 & 7.0 & 1.3 & 74.6\\
    \tikz\fill[ceiling] (0,0) circle (3.5pt);\:GFNet-XS~\cite{rao2021global} & 2021 & 16.0 & 2.9 & 78.6\\
    \tikz\fill[ceiling] (0,0) circle (3.5pt);\:GFNet-S~\cite{rao2021global} & 2021 & 25.0 & 4.5 & 80.0\\
    \tikz\fill[ceiling] (0,0) circle (3.5pt);\:GFNet-B~\cite{rao2021global} & 2021 & 43.0 & 7.9 & 80.7\\
    \hline
    \rowcolor{gray!20} \tikz\fill[objects] (0,0) circle (3.5pt);\:Focal-Tiny~\cite{yang2021focal} & 2021 & 29.1 & 4.9 & 82.2\\
    \rowcolor{gray!20} \tikz\fill[objects] (0,0) circle (3.5pt);\:Focal-Small~\cite{yang2021focal} & 2021 & 51.1 & 9.1 & 83.5\\
    \rowcolor{gray!20} \tikz\fill[objects] (0,0) circle (3.5pt);\:Focal-Base~\cite{yang2021focal} & 2021 & 89.8 & 16.0 & 83.8\\
    \hline
    \tikz\fill[trunk] (0,0) circle (3.5pt);\:T2T-ViT-14~\cite{Yuan_2021_ICCV} & 2021 & 21.5 & 4.8 & 81.5\\
    \tikz\fill[trunk] (0,0) circle (3.5pt);\:T2T-ViT-19~\cite{Yuan_2021_ICCV} & 2021 & 39.2 & 8.5 & 81.9\\
    \tikz\fill[trunk] (0,0) circle (3.5pt);\:T2T-ViT-24~\cite{Yuan_2021_ICCV} & 2021 & 64.1 & 13.8 & 82.3\\
    \hline
    \rowcolor{gray!20} \tikz\fill[sofa] (0,0) -- (8pt,0) -- (4pt,8pt) -- cycle;\:CSWin-T~\cite{dong2022cswin} & 2022 & 23.0 & 4.3 & 82.7\\
    \rowcolor{gray!20} \tikz\fill[sofa] (0,0) -- (8pt,0) -- (4pt,8pt) -- cycle;\:CSWin-S~\cite{dong2022cswin} & 2022 & 35.0 & 6.9 & 83.6\\
    \rowcolor{gray!20} \tikz\fill[sofa] (0,0) -- (8pt,0) -- (4pt,8pt) -- cycle;\:CSWin-B~\cite{dong2022cswin} & 2022 & 78.0 & 15.0 & 84.2\\
    \hline
    \tikz\fill[parking](0,0) -- (8pt,0) -- (4pt,8pt) -- cycle;\:MaxSA-T~\cite{tu2022maxvit} & 2022 & 31.0 & 5.6 & 83.6\\
    \tikz\fill[parking](0,0) -- (8pt,0) -- (4pt,8pt) -- cycle;\:MaxSA-S~\cite{tu2022maxvit} & 2022 & 69.0 & 11.7 & 84.5\\
    \tikz\fill[parking](0,0) -- (8pt,0) -- (4pt,8pt) -- cycle;\:MaxSA-B~\cite{tu2022maxvit} & 2022 & 120.0 & 23.4 & 85.0\\
    \tikz\fill[parking](0,0) -- (8pt,0) -- (4pt,8pt) -- cycle;\:MaxSA-L~\cite{tu2022maxvit} & 2022 & 212.0 & 43.9 & 85.2\\
        
    \bottomrule[1.5px]
    \end{tabular}
        }
    \end{minipage}
    \hfill
     \begin{minipage}[[t]{0.49\linewidth}
     \renewcommand{\arraystretch}{1.44} 
     \centering
     \resizebox{\linewidth}{!}{
        \begin{tabular}{lcrrr}
        \bottomrule[1.5px]
         \textbf{Model} &\textbf{Year} & \textbf{\makecell[r]{Params \\ (M)$\downarrow$}}  & \textbf{\makecell[r]{FLOPs \\ (G)$\downarrow$}}  & \textbf{\makecell[r]{Top-1 Acc. \\ (\%)$\uparrow$}}\\
    \hline
        \rowcolor{gray!20} \tikz\fill[other_veh] (0,0) rectangle (8pt,6pt);\:{SKIPAT$|$ViT-T}~\cite{venkataramanan2023skipattention} & 2023 & 5.8 & 1.1 & 72.9\\
    \rowcolor{gray!20} \tikz\fill[other_veh] (0,0) rectangle (8pt,6pt);\:{SKIPAT$|$ViT-S}~\cite{venkataramanan2023skipattention} & 2023 & 22.1 & 4.0 & 80.2\\
    \rowcolor{gray!20} \tikz\fill[other_veh] (0,0) rectangle (8pt,6pt);\:{SKIPAT-B$|$ViT-B}~\cite{venkataramanan2023skipattention} & 2023 & 86.7 & 15.2 & 82.2\\   
    \hline
    \tikz\fill[trunk] (0,0) rectangle (8pt,6pt);\:SwiftFormer-XS~\cite{shaker2023swiftformer} & 2023 & 3.5 & - & 75.7\\
    \tikz\fill[trunk] (0,0) rectangle (8pt,6pt);\:SwiftFormer-S~\cite{shaker2023swiftformer} & 2023 & 6.1 & - & 78.5\\
     \tikz\fill[trunk] (0,0) rectangle (8pt,6pt);\:SwiftFormer-L1~\cite{shaker2023swiftformer} & 2023 & 12.1 & - & 80.9\\
      \hline
    \rowcolor{gray!20} \tikz\fill[parking] (0,0) rectangle (8pt,6pt);\:BiFormer-T~\cite{zhu2023biformer} & 2023 & 13.1 & 2.2 & 81.4\\
    \rowcolor{gray!20} \tikz\fill[parking] (0,0) rectangle (8pt,6pt);\:BiFormer-S~\cite{zhu2023biformer} & 2023 & 26.0 & 4.5 & 83.8\\
    \rowcolor{gray!20} \tikz\fill[parking] (0,0) rectangle (8pt,6pt);\:BiFormer-B~\cite{zhu2023biformer} & 2023 & 57.0 & 9.8 & 84.3\\
     \hline
    \tikz\fill[truck] (0,0) rectangle  (6.5pt,6.5pt);\:REMViTv2-T~\cite{diko2024revit} & 2024 & 24.0 & 4.7 & 82.7\\
    \tikz\fill[truck] (0,0) rectangle  (6.5pt,6.5pt);\:REViT-T~\cite{diko2024revit} & 2024 & 29.0 & 4.5 & 81.5\\
    \tikz\fill[truck] (0,0) rectangle  (6.5pt,6.5pt);\:RESwim-T~\cite{diko2024revit} & 2024 & 29.0 & 4.5 & 81.5\\
    \tikz\fill[truck] (0,0) rectangle  (6.5pt,6.5pt);\:REViT-B~\cite{diko2024revit} & 2024 & 86.0 & 17.5 & 82.4\\
    \hline
    \rowcolor{gray!20} \tikz\fill[pole] (0,0) rectangle (6.5pt,6.5pt);\:SSViT-T~\cite{fan2024visiontransformersparsescan} & 2024 & 15.0 &1.2&83.0\\
    \rowcolor{gray!20} \tikz\fill[pole] (0,0) rectangle (6.5pt,6.5pt);\:SSViT-S~\cite{fan2024visiontransformersparsescan} & 2024 & 27.0 &2.2&84.4\\
    \rowcolor{gray!20} \tikz\fill[pole] (0,0) rectangle (6.5pt,6.5pt);\:SSViT-B~\cite{fan2024visiontransformersparsescan} & 2024 & 57.0 &4.8&\underline{85.3}\\
    \rowcolor{gray!20} \tikz\fill[pole] (0,0) rectangle (6.5pt,6.5pt);\:SSViT-L~\cite{fan2024visiontransformersparsescan} & 2024 & 100.0 &9.1& \textbf{85.7}\\
    \bottomrule[1.5px]
        \end{tabular}
        }
    \end{minipage}
    \label{table:self attention}
\end{table*}

\noindent 
\textbf{{Partial Attention~(PA).}}  
PA focuses on limiting the scope of attention to a subset of tokens, thereby reducing computational demands. The Swin Transformer\footnote{\url{https://github.com/microsoft/Swin-Transformer}}~\citep{liu2021swin} achieves this by using shifted windows to confine self-attention computations to non-overlapping local windows, resulting in linear computational growth. Specifically, assuming each window contains $ M \times M $ pixels, the computational complexities of the standard global Multi-Head Self-Attention (MHSA) module and \  the Window-based Multi-head Self-Attention (WMSA) on the input image with $ h \times w $ patches are as follows:
\begin{equation}
    \begin{aligned}
        \Omega(\mathrm{MSA}) &= 4 h w \mathrm{C}^{2} + 2(h w)^{2} \mathrm{C} \\
\Omega(\mathrm{W}\mathrm{MSA}) &= 4 h w \mathrm{C}^{2} + 2 M^{2} h w \mathrm{C}
    \end{aligned}
\end{equation}
where the $\mathrm{C}$ denotes the embedding dimension,  symbol $\Omega$ is used to measure the computation complexity, $\Omega(\mathrm{MHSA})$ has a quadratic complexity with respect to the number of patches $h \times w$, while the $\Omega(\mathrm{W}\mathrm{MSA})$ has a promising linear complexity.
Similarly, the CSWin Transformer~\footnote{\url{https://github.com/microsoft/CSWin-Transformer}}~\citep{dong2022cswin} partitions the input features into horizontal and vertical stripes and computes SA within these localized regions, as shown in Figure~\ref{fig:cwwin}. 
The TNT~\footnote{\url{https://github.com/lucidrains/transformer-in-transformer}} model \citep{Han2021TransformerIT} employs transformers at the patch and pixel levels, facilitating detailed attention to local features.

\noindent 
\textbf{{Local Self-Attention.}} 
Local SA takes the idea of partial attention by restricting attention operations to local regions of the input image~\citep{Zhou2023-gl}. 
Multi-axis attention (Max-SA) \citep{tu2022maxvit} simplifies the complexity of attention from quadratic to linear by implementing windowed and mesh attention, while ReViT \citep{diko2024revit} addresses feature collapse in deep ViT layers through residual attention learning. Deformable self-attention \citep{xia2022vision} and dynamic sparse attention \citep{zhu2023biformer,Tamura2024-ei} dynamically select attention blocks based on data dependencies and query patterns, respectively. The Focal Self-attention model \citep{yang2021focal} captures both short-range and long-range dependencies, utilizing fine-grained local attention and coarse-grained distant interactions. ViL\citep{zhang2021multi} combines multi-scale architectures with extended context considerations, improving high-resolution image processing and encoding.

\begin{figure*}
    \centering
    \includegraphics[width=\textwidth]{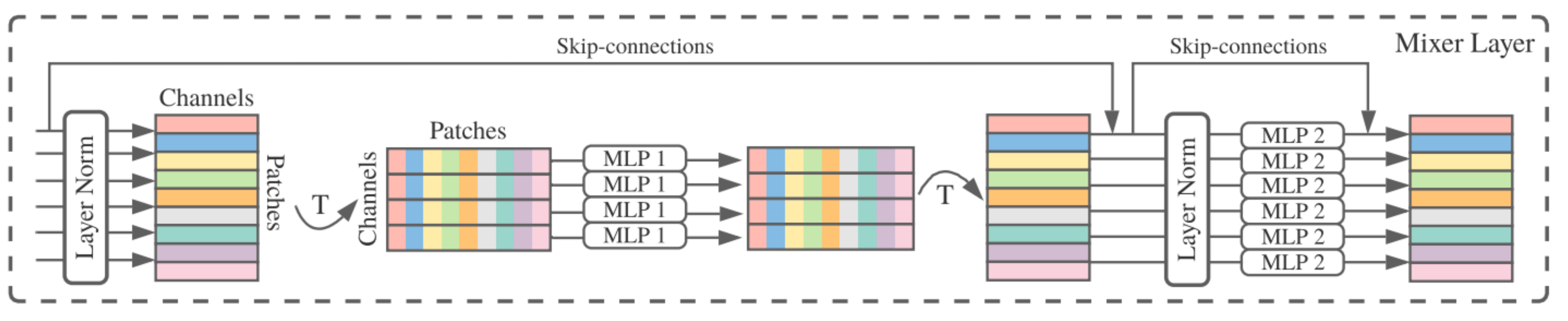}
    \caption{
    \textbf{Attention-Free Paradigm for Token Updates.}
    MLP-Mixer~\citep{tolstikhin2021mlpmixer} proposes to replace convolution and attention in ViT with MLP models. The figure shows the key part of the MLP-Mixer, which contains mixing patches and mixing channels.
    }
    \label{fig:mlp}
\end{figure*}

\noindent 
\textbf{{Linear Self-Attention.}} 
Linear SA aims to reduce the computational complexity of SA from quadratic to linear~\citep{Lu2024-kg}.
Various methods are proposed, such as replacing dot product attention with position-sensitive hashing \citep{kitaev2020reformer}.
GFNet~\citep{rao2021global} employs discrete two-dimensional Fourier transforms combined with a learnable global filter, and Linformer~\citep{wang2020linformer} introduces a low-rank matrix to achieve linear complexity.
SwiftFormer \citep{shaker2023swiftformer} uses an efficient additive attention mechanism, while \citep{mehta2022separable} proposes separable self-attention using element-wise intelligent operations. The neighborhood attention mechanism \citep{sun2023vicinity} adjusts attention weights based on the 2D Manhattan distance from neighboring patches. 
Similarly, Vision Outlooker (VOLO) \citep{9888055}, while primarily focusing on computer vision tasks, introduces an innovative locality-sensitive attention mechanism, balancing efficiency and accuracy.

\noindent 
\textbf{{Low-Rank Decomposition.}} 
These techniques decompose the attention matrix into low-dimensional representations to optimize the computation. MobileViT \citep{mobilevit2021} combines low-rank decomposition and sparsification techniques to transform ViT into a mobile-friendly version. The Low-Rank Transformer \citep{lowranktransformer2020} reduces computational complexity by decomposing the SA layer, while the Performer~\citep{performer2020} integrates kernel techniques with low-rank decomposition.
Tensor decomposition techniques, such as CP-Decomposition \citep{tensordecomposition2020, cpdecomposition2014}, and factorized self-attention mechanisms \citep{factorizedselfattention2021} have also been applied for model optimization.

\subsubsection{\textbf{Attention-Free Architecture}}
\label{Attention-Free Architecture}
While self-attention revolutionizes the field of image recognition, its computational complexity motivates the exploration of alternative architectures that achieve comparable performance with reduced resource requirements and leads to renewed interest in MLP, a classic neural network architecture consisting of fully connected layers and activation functions.

MLP-based architectures, often dispensing with the complexities of convolutional and self-attention mechanisms, offer advantages in terms of adaptability and computational efficiency. 
This topic focuses on two main directions in MLP research for efficient image recognition: \textbf{Pure MLP Architecture} and \textbf{Hybrid MLP-CNN Architecture}. 
These approaches highlight the potential of the MLP design as a competitive alternative to attention-based models.

\noindent 
\textbf{{Pure MLP Architecture.}}
Pure MLP architectures aim to simplify the token update operation for improved efficiency while maintaining performance. 
ResMLP~\citep{touvron2021resmlp} introduces a residual network architecture that alternates between linear layers and two-layer feedforward networks. 
As shown in Figure~\ref{fig:mlp}, MLP-Mixer\footnote{\url{https://github.com/jaketae/mlp-mixer}}~\citep{tolstikhin2021mlpmixer} introduces Token-Mixing and Channel-Mixing MLP blocks to process images, avoiding convolutions and SA. 
S2-MLP~\citep{yu2021s2mlp}, inspired by MLP-Mixer, relies on channel mixing and incorporates a spatial shift operation for patch interaction. 
AS-MLP\footnote{\url{https://github.com/svip-lab/AS-MLP}}~\citep{lian2022asmlp} adopts axially shifted channels to capture local dependencies, allowing the pure MLP model to simulate the local receptive fields of CNN models. 
MLP-Unet~\citep{10138195} applies a fully MLP-based method and cascaded upsampler to keep local information effectively.

\begin{table*}[tb]
\caption{
A quantitative comparative analysis is conducted on the classic methods involved in \textbf{Attention-Free Design}~(Section~\ref{Attention-Free Architecture}), using ImageNet-1K as the benchmark, with the resolution standardized to $224 \times 224$.
The “/16” in Mixer means the model of base scale with patches of resolution $16 \times 16$
The best-performing method is highlighted in bold, and the runner-up is underlined.
}
    \centering
    \begin{minipage}[t]{0.49\linewidth}
    \renewcommand{\arraystretch}{1.36} 
    \resizebox{\linewidth}{!}{
        \begin{tabular}{lccrr}
        \bottomrule[1.5px]
         \textbf{Model} &\textbf{Year} & \textbf{\makecell[r]{Params \\ (M)$\downarrow$}}  & \textbf{\makecell[r]{FLOPs \\ (G)$\downarrow$}}  & \textbf{\makecell[r]{Top-1 Acc. \\ (\%)$\uparrow$}}\\
    \hline
    \tikz\fill[gray] (0,0) -- (8pt,0) -- (4pt,8pt) -- cycle;\:GMLP-T~\cite{liu2021pay} & 2021 & 6.0 & - & 72.3\\
    \tikz\fill[gray] (0,0) -- (8pt,0) -- (4pt,8pt) -- cycle;\:GMLP-S~\cite{liu2021pay} & 2021 & 20.0 & - & 79.6\\
    \tikz\fill[gray] (0,0) -- (8pt,0) -- (4pt,8pt) -- cycle;\:GMLP-B~\cite{liu2021pay} & 2021 & 73.0 & - & 81.6\\
    \hline
    \rowcolor{gray!20} \tikz\fill[trunk] (0,0) -- (8pt,0) -- (4pt,8pt) -- cycle;\:HireMLP-Tiny~\cite{guo2021hiremlp} & 2021 & 18.0 & 2.1 & 79.7\\
    \rowcolor{gray!20} \tikz\fill[trunk] (0,0) -- (8pt,0) -- (4pt,8pt) -- cycle;\:HireMLP-Small~\cite{guo2021hiremlp} & 2021 & 33.0 & 4.2 & 82.1\\
    \rowcolor{gray!20} \tikz\fill[trunk] (0,0) -- (8pt,0) -- (4pt,8pt) -- cycle;\:HireMLP-Base~\cite{guo2021hiremlp} & 2021 & 58.0 & 8.1 & 83.2\\
    \rowcolor{gray!20} \tikz\fill[trunk] (0,0) -- (8pt,0) -- (4pt,8pt) -- cycle;\:HireMLP-Large~\cite{guo2021hiremlp} & 2021 & 96.0 & 13.4 & 83.8\\
    \hline
    \tikz\fill[tvs] (0,0) -- (8pt,0) -- (4pt,8pt) -- cycle;\:Mixer-B/16~\cite{tolstikhin2021mlpmixer} & 2021 & 59.0 & 11.6 & 82.4 \\
    \tikz\fill[tvs] (0,0) -- (8pt,0) -- (4pt,8pt) -- cycle;\:Mixer-L/16~\cite{tolstikhin2021mlpmixer} & 2021 & 207.0 & - & 77.1\\
   \hline
   \rowcolor{gray!20} \tikz\fill[sofa] (0,0) -- (8pt,0) -- (4pt,8pt) -- cycle;\:ConvMLP-S~\cite{li2021convmlp} & 2021 & 9.0 & 2.4 & 76.8\\
   \rowcolor{gray!20}  \tikz\fill[sofa] (0,0) -- (8pt,0) -- (4pt,8pt) -- cycle;\:ConvMLP-M~\cite{li2021convmlp} & 2021 & 17.4 & 3.9 & 79.0\\
    \rowcolor{gray!20} \tikz\fill[sofa] (0,0) -- (8pt,0) -- (4pt,8pt) -- cycle;\:ConvMLP-L~\cite{li2021convmlp} & 2021 & 42.7 & 9.9 & 80.2\\
    \hline
    \tikz\fill[trunk] (0,0) rectangle (8pt,6pt);\:AS-MLP-T~\cite{lian2022asmlp} & 2022 & 28.0 & 4.4 & 81.3\\
    \tikz\fill[trunk] (0,0) rectangle (8pt,6pt);\:AS-MLP-S~\cite{lian2022asmlp} & 2022 & 50.0 & 8.5 & 83.1\\
    \tikz\fill[trunk] (0,0) rectangle (8pt,6pt);\:AS-MLP-B~\cite{lian2022asmlp} & 2022 & 88.0 & 15.2 & 83.3\\
    \hline
    \rowcolor{gray!20} \tikz\fill[gray] (0,0) rectangle  (8pt,6pt);\:ResMLP-S12~\cite{touvron2022resmlp} & 2022 & 15.0 & 3.0 & 76.6\\
    \rowcolor{gray!20} \tikz\fill[gray] (0,0) rectangle  (8pt,6pt);\:ResMLP-S24~\cite{touvron2022resmlp} & 2022 & 30.0 & 6.0 & 79.4\\
    \rowcolor{gray!20} \tikz\fill[gray] (0,0) rectangle  (8pt,6pt);\:ResMLP-B24~\cite{touvron2022resmlp} & 2022 & 116.0 & 23.0 & 81.0\\
    \bottomrule[1.5px]
    \end{tabular}
        }
    \end{minipage}
    \hfill
    \centering
     \begin{minipage}[[t]{0.49\linewidth}
     \renewcommand{\arraystretch}{1.39} 
     \resizebox{\linewidth}{!}{
        \begin{tabular}{lccrr}
        \bottomrule[1.5px]
         \textbf{Model} &\textbf{Year} & \textbf{\makecell[r]{Params \\ (M)$\downarrow$}}  & \textbf{\makecell[r]{FLOPs \\ (G)$\downarrow$}}  & \textbf{\makecell[r]{Top-1 Acc. \\ (\%)$\uparrow$}}\\
    \hline
   
    \tikz\fill[sofa] (0,0) rectangle  (8pt,6pt);\:CycleMLP-B1~\cite{chen2022cyclemlp} & 2022 & 15.0 & 2.1 & 79.1\\
    \tikz\fill[sofa] (0,0) rectangle  (8pt,6pt);\:CycleMLP-B2~\cite{chen2022cyclemlp} & 2022 & 27.0 & 3.9 & 81.6\\
    \tikz\fill[sofa] (0,0) rectangle  (8pt,6pt);\:CycleMLP-T~\cite{chen2022cyclemlp} & 2022 & 28.0 & 4.4 & 81.3\\
    \tikz\fill[sofa] (0,0) rectangle  (8pt,6pt);\:CycleMLP-B3~\cite{chen2022cyclemlp} & 2022 & 38.0 & 6.9 & 82.4\\
    \tikz\fill[sofa] (0,0) rectangle  (8pt,6pt);\:CycleMLP-S~\cite{chen2022cyclemlp} & 2022 & 50.0 & 8.5 & 82.9\\
    \tikz\fill[sofa] (0,0) rectangle  (8pt,6pt);\:CycleMLP-B~\cite{chen2022cyclemlp} & 2022 & 88.0 & 15.2 & 83.4\\
    \hline
    \rowcolor{gray!20} \tikz\fill[objects] (0,0) rectangle  (8pt,6pt);\:S$^2$-MLP-deep~\cite{yu2022s2} & 2022 & 51.0 & 10.5 & 80.7\\
    \rowcolor{gray!20} \tikz\fill[objects] (0,0) rectangle  (8pt,6pt);\:S$^2$-MLP-wide~\cite{yu2022s2} & 2022 & 71.0 & 14.0 & 80.0\\
    \hline
    \tikz\fill[tvs] (0,0) rectangle  (8pt,6pt);\:PoolFormer-S12~\cite{yu2022metaformer} & 2022 & 12.0 & 1.8 & 77.2\\
    \tikz\fill[tvs] (0,0) rectangle  (8pt,6pt);\:PoolFormer-S24~\cite{yu2022metaformer} & 2022 & 21.0 & 3.4 & 80.3\\
    \tikz\fill[tvs] (0,0) rectangle  (8pt,6pt);\:PoolFormer-S36~\cite{yu2022metaformer} & 2022 & 31.0 & 5.0 & 81.4\\
    \tikz\fill[tvs] (0,0) rectangle  (8pt,6pt);\:PoolFormer-M36~\cite{yu2022metaformer} & 2022 & 56.0 & 8.8 & 82.1\\
    \tikz\fill[tvs] (0,0) rectangle  (8pt,6pt);\:PoolFormer-M48~\cite{yu2022metaformer} & 2022 & 73.0 & - & 82.5\\
    \hline
    \rowcolor{gray!20} \tikz\fill[pole] (0,0) circle (3.5pt);\:SVT-H-S~\cite{patro2024scattering} & 2023&22.0 &3.9 & 84.2\\
    \rowcolor{gray!20} \tikz\fill[pole] (0,0) circle (3.5pt);\:SVT-H-B~\cite{patro2024scattering} & 2023&32.8 &6.3 & \underline{85.2} \\
    \rowcolor{gray!20} \tikz\fill[pole] (0,0)circle (3.5pt);\:SVT-H-L~\cite{patro2024scattering} & 2023&54.0 &12.7 & \textbf{85.7}\\
    \hline
    \tikz\fill[truck] (0,0) rectangle  (6.5pt,6.5pt);\:MONet-T~\cite{cheng2024multilinearoperatornetworks} & 2024 & 10.3 & 2.8 &77.0\\
    \tikz\fill[truck] (0,0) rectangle  (6.5pt,6.5pt);\:MONet-S~\cite{cheng2024multilinearoperatornetworks} & 2024 & 32.9 & 6.8 &81.3\\
    \bottomrule[1.5px]
        \end{tabular}
        }
    \end{minipage}
    \label{table:MLP}
    \vspace{-0.4cm}
\end{table*}

\noindent 
\textbf{{Hybrid MLP-CNN Structure.}}
Hybrid attention-free structures that integrate MLP and CNN offer a promising balance of complementary strengths. 
X-MLP~\citep{wang2023xmlp} employs linear projections for feature map interactions, eliminating the need for convolution operations in patch embedding. \citep{wang2023pretrainingattention} propose a pre-training method that does not rely on attention mechanisms. gMLP~\citep{liu2021pay} identifies the high computational burden of self-attention and uses channel and static spatial projections to maintain high precision with fewer parameters. 
A hybrid model~\citep{Ji2023RMMLPRollingMA} achieves performance comparable to CNN-Transformer~\citep{Cao2024-rm} hybrids by employing matrix decomposition and multiscale information fusion. 
\citep{gu2022efficientlymodelinglongsequences} propose a structured state space method for efficiently modeling long sequences. US-MLP~\citep{Luo2023USMLPUS} integrates CNN layers with sparse MLP blocks. 
Furthermore, GSS model~\citep{mehta2022longrangelanguagemodeling} effectively processes long sequence modeling through a gated state space, providing immense potential for image recognition applications. Similarly, \citep{gu2022efficientlymodelinglongsequences} addresses the computational bottleneck in long sequence processing by optimizing state space modeling. PoolFormer~\citep{yu2022metaformer} \footnote{ \url{https://github.com/sail-sg/poolformer} }replaces self-attention with grouping operations as a token mixer.
Finally, Mamba~\citep{gu2024mambalineartimesequencemodeling} achieves linear time complexity in sequence modeling through selective state space. By applying a selective state space, Mamba effectively balances accuracy and computational burden in sequence modeling~\citep{zhu2024visionmambaefficientvisual,yao2024spectralmambaefficientmambahyperspectral,pei2024efficientvmambaatrousselectivescan,chen2024mimistdmambainmambaefficientinfrared,Tang_2024_CVPR}.

\subsubsection{\textbf{Summary: Efficient Token Update}}

\noindent
\textbf{Self-Attention Optimization.}
While self-attention captures long-distance dependencies, its computational complexity increases with large datasets like ImageNet-1K. The goal is to optimize complexity without losing global dependency capture.
We report the relevant experimental results in Table ~\ref{table:self attention}, where it can be observed that:
\textit{\textbf{(i).}} Methods based on local self-attention, such as GFNet~\citep{rao2023gfnet} and SKIPAT~\citep{venkataramanan2023skipattention}, slightly compromise accuracy but significantly reduce computational burden. For example, SKIPAT~\citep{venkataramanan2023skipattention} reduces the parameters to 5.8M and the FLOPs to $1.1$G;
\textit{\textbf{(ii).}} BiFormer~\citep{zhu2023biformer} achieves an excellent balance between parameters, computational cost, and accuracy, with 13.1M parameters and only $2.2$G FLOPs, while attaining an accuracy of $81.4\%$;
\textit{\textbf{(iii).}} SSViT~\citep{fan2024visiontransformersparsescan}, as a state-of-the-art method, demonstrates an impressive accuracy of $85.7\%$ with only $9.1$G FLOPs, which is lower than the DeiT standard model, indicating that optimization of self-attention still holds potential.

\begin{table*}[tb]
\caption{
A quantitative comparative analysis is conducted on the classic methods involved in \textbf{Efficient Framework Design}~(Section~\ref{Framework Design}), using ImageNet-1K as the benchmark, with the resolution standardized to $224 \times 224$.
The best-performing method is highlighted in bold, and the runner-up is underlined.
}
    \begin{minipage}[t]{0.49\linewidth}
    \renewcommand{\arraystretch}{1.36} 
    \centering
    \resizebox{\linewidth}{!}{
        \begin{tabular}{lcrrr}
        \bottomrule[1.5px]
         \textbf{Model} &\textbf{Year} & \textbf{\makecell[r]{Params \\ (M)$\downarrow$}}  & \textbf{\makecell[r]{FLOPs \\ (G)$\downarrow$}}  & \textbf{\makecell[r]{Top-1 Acc. \\ (\%)$\uparrow$}}\\
    \hline
    \tikz\fill[gray] (0,0) -- (8pt,0) -- (4pt,8pt) -- cycle;\:PVT-Tiny~\cite{wang2021pyramid} & 2021 & 13.2 & 1.9 & 75.1\\
    \tikz\fill[gray] (0,0) -- (8pt,0) -- (4pt,8pt) -- cycle;\:PVT-Small~\cite{wang2021pyramid} & 2021 & 24.5 & 3.8 & 79.8\\
    \tikz\fill[gray] (0,0) -- (8pt,0) -- (4pt,8pt) -- cycle;\:PVT-Medium~\cite{wang2021pyramid} & 2021 & 44.2 & 6.7 & 81.2\\
    \tikz\fill[gray] (0,0) -- (8pt,0) -- (4pt,8pt) -- cycle;\:PVT-Large~\cite{wang2021pyramid} & 2021 & 61.4 & 9.8 & 81.7\\
    \hline
    \rowcolor{gray!20} \tikz\fill[trunk] (0,0) -- (8pt,0) -- (4pt,8pt) -- cycle;\:HAT-Net-Tiny~\cite{liu2024visiontransformershierarchicalattention} & 2021 & 12.7 & 2.0 & 79.8\\
    \rowcolor{gray!20} \tikz\fill[trunk] (0,0) -- (8pt,0) -- (4pt,8pt) -- cycle;\:HAT-Net-Small~\cite{liu2024visiontransformershierarchicalattention} & 2021 & 25.7 & 4.3 & 82.6\\
    \rowcolor{gray!20} \tikz\fill[trunk] (0,0) -- (8pt,0) -- (4pt,8pt) -- cycle;\:HAT-Net-Medium~\cite{liu2024visiontransformershierarchicalattention} & 2021 & 42.9 & 8.3 & 84.0\\
    \rowcolor{gray!20} \tikz\fill[trunk] (0,0) -- (8pt,0) -- (4pt,8pt) -- cycle;\:HAT-Net-Large~\cite{liu2024visiontransformershierarchicalattention} & 2021 & 63.1 & 11.5 & 84.2\\
    \hline
    \tikz\fill[tvs] (0,0) -- (8pt,0) -- (4pt,8pt) -- cycle;\:Twims-SVT-S~\cite{chu2021twins} & 2021 & 24.0 & 2.9 & 81.7\\
    \tikz\fill[tvs] (0,0) -- (8pt,0) -- (4pt,8pt) -- cycle;\:Twims-SVT-S~\cite{chu2021twins} & 2021 & 24.1 & 3.8 & 81.2\\
    \tikz\fill[tvs] (0,0) -- (8pt,0) -- (4pt,8pt) -- cycle;\:Twims-SVT-B~\cite{chu2021twins} & 2021 & 56.0 & 8.6 & 83.2\\
    \tikz\fill[tvs] (0,0) -- (8pt,0) -- (4pt,8pt) -- cycle;\:Twims-PCPVT-B~\cite{chu2021twins} & 2021 & 43.8 & 6.7 & 82.7\\
    \tikz\fill[tvs] (0,0) -- (8pt,0) -- (4pt,8pt) -- cycle;\:Twims-PCPVT-L~\cite{chu2021twins} & 2021 & 60.9 & 9.8 & 83.1\\
   \tikz\fill[tvs] (0,0) -- (8pt,0) -- (4pt,8pt) -- cycle;\:Twims-SVT-L~\cite{chu2021twins} & 2021 & 99.2 & 15.1 & 83.7\\
   \hline
   \rowcolor{gray!20} \tikz\fill[sofa] (0,0) -- (8pt,0) -- (4pt,8pt) -- cycle;\:Swim-T~\cite{liu2021swin} & 2021 & 29.0 & 4.5 & 81.3\\
    \rowcolor{gray!20} \tikz\fill[sofa] (0,0) -- (8pt,0) -- (4pt,8pt) -- cycle;\:Swim-S~\cite{liu2021swin} & 2021 & 50.0 & 8.7 & 83.0\\
    \rowcolor{gray!20} \tikz\fill[sofa] (0,0) -- (8pt,0) -- (4pt,8pt) -- cycle;\:Swim-B~\cite{liu2021swin} & 2021 & 88.0 & 15.4 & 83.5\\
   \hline
    \tikz\fill[objects] (0,0) circle (3.5pt);\:MaxViT-T~\cite{tu2022maxvit} & 2022 & 31.0 & 5.6 & 83.6\\
    \tikz\fill[objects] (0,0) circle (3.5pt);\:MaxViT-S~\cite{tu2022maxvit} & 2022 & 69.0 & 11.7 & 84.5\\
    \bottomrule[1.5px]
    \end{tabular}
        }
    \end{minipage}
    \hfill
     \begin{minipage}[[t]{0.49\linewidth}
     \renewcommand{\arraystretch}{1.36} 
     \centering
     \resizebox{\linewidth}{!}{
        \begin{tabular}{lccrr}
        \bottomrule[1.5px]
          \textbf{Model} &\textbf{Year} & \textbf{\makecell[r]{Params \\ (M)$\downarrow$}}  & \textbf{\makecell[r]{FLOPs \\ (G)$\downarrow$}}  & \textbf{\makecell[r]{Top-1 Acc. \\ (\%)$\uparrow$}}\\
    \hline
    \tikz\fill[objects] (0,0) circle (3.5pt);\:MaxViT-B~\cite{tu2022maxvit} & 2022 & 120.0 & 23.4 & 85.0\\
    \tikz\fill[objects] (0,0) circle (3.5pt);\:MaxViT-L~\cite{tu2022maxvit} & 2022 & 212.0 & 43.9 & 85.2\\
    \hline
    \rowcolor{gray!20} \tikz\fill[trunk] (0,0) rectangle (8pt,6pt);\:FDViT-Ti~\cite{inproceedings} & 2023 & 4.5 & 0.6 & 73.7\\
    \rowcolor{gray!20} \tikz\fill[trunk] (0,0) rectangle (8pt,6pt);\:FDViT-S~\cite{xu2023fdvit} & 2023 & 21.5 & 2.8 & 81.5\\
    \rowcolor{gray!20} \tikz\fill[trunk] (0,0) rectangle (8pt,6pt);\:FDViT-B~\cite{xu2023fdvit} & 2023 & 67.8 & 11.9 & 82.4\\
    \hline
    \tikz\fill[gray] (0,0) rectangle  (8pt,6pt);\:HiViT-T~\cite{zhang2023hivit} & 2023 & 19.2 & 4.6 & 82.1\\
    \tikz\fill[gray] (0,0) rectangle  (8pt,6pt);\:HiViT-S~\cite{zhang2023hivit} & 2023 & 38.5 & 9.1 & 83.5\\
    \tikz\fill[gray] (0,0) rectangle  (8pt,6pt);\:HiViT-L~\cite{zhang2023hivit} & 2023 & 66.4 & 15.9 & 83.8\\
     \hline
    \rowcolor{gray!20} \tikz\fill[sofa] (0,0) rectangle  (8pt,6pt);\:Hiera-T~\cite{ryali2023hiera} & 2023 & 28.0 & 5.0 & 82.8\\
    \rowcolor{gray!20} \tikz\fill[sofa] (0,0) rectangle  (8pt,6pt);\:Hiera-S~\cite{ryali2023hiera} & 2023 & 35.0 & 6.0 & 83.8\\
    \rowcolor{gray!20} \tikz\fill[sofa] (0,0) rectangle  (8pt,6pt);\:Hiera-B~\cite{ryali2023hiera} & 2023 & 52.0 & 9.0 & 84.5 \\
    \rowcolor{gray!20} \tikz\fill[sofa] (0,0) rectangle  (8pt,6pt);\:Hiera-B+~\cite{ryali2023hiera} & 2023 & 70.0 & 13.0 & 85.2 \\
    \rowcolor{gray!20} \tikz\fill[sofa] (0,0) rectangle  (8pt,6pt);\:Hiera-L~\cite{ryali2023hiera} & 2023 & 214.0 & 40.0 & \underline{86.1} \\
    \rowcolor{gray!20} \tikz\fill[sofa] (0,0) rectangle  (8pt,6pt);\:Hiera-H~\cite{ryali2023hiera} & 2023 & 673.0 & 125.0 & \textbf{86.9}\\
    \hline
    \tikz\fill[bed] (0,0) rectangle  (6.5pt,6.5pt);\:HIRI-ViT-S~\cite{10475592} & 2024 & 34.8 & 4.5 & 83.6\\
    \tikz\fill[bed] (0,0) rectangle  (6.5pt,6.5pt);\:HIRI-ViT-B~\cite{10475592} & 2024 & 54.4 & 8.2 & 84.7\\
    \tikz\fill[bed] (0,0) rectangle  (6.5pt,6.5pt);\:HIRI-ViT-L~\cite{10475592} & 2024 & 94.4 & 17.0 & 85.3\\
    \bottomrule[1.5px]
        \end{tabular}
        }
    \end{minipage}
    \label{table:hierarchical structures}
\end{table*}

\noindent
\textbf{Attention-Free Design.}
The MLP design enhances computational efficiency by eliminating complex attention calculations, enabling faster adaptation to various hardware and optimization strategies.
When combined with CNNs, MLPs leverage their computational efficiency while keeping the exceptional feature extraction capabilities of CNNs, achieving a balance between performance and efficiency. 
Table Y presents comparative results of relevant methods, revealing the following observations:
\textit{\textbf{(i).}} HireMLP~\citep{guo2021hiremlp} series exhibit outstanding performance with relatively fewer parameters and lower FLOPs. For example, HireMLP-Small achieves an accuracy of $82.1\%$ using only 33M parameters and 4.2G FLOPs.
\textit{\textbf{(ii).}} The hybrid MLP-CNN architecture can be adjusted according to specific requirements, accommodating various application scenarios and resource constraints. HireMLP-Small~\citep{guo2021hiremlp} achieves an accuracy of $82.1\%$ with only $4.2$ G FLOPs.
\textit{\textbf{(iii).}} Particularly notable is SVT-H-L~\citep{sun2023vicinity}, which achieves an impressive Top-1 accuracy of $85.7\%$ using only 54M parameters and $12.7$G FLOPs.

\subsection{\textbf{Framework Design}}
\label{Framework Design}
Inspired by hierarchical structures in convolutional neural networks for feature extraction and image pyramiding, researchers are integrating these principles into vision transformer (ViT) architectures. This hierarchical design aims to improve the efficiency of ViT and the learning of representation. 
Hierarchical ViTs differentiate background and foreground objects, improving image recognition in complex scenes with varied scales or occlusions~\citep{Wang2024-eu}. 
This topic explores hierarchical ViT design, with the aim of capturing features at multiple scales for both global context and fine details.


\noindent 
\textbf{{Hierarchical Framework.}}
Swin Transformer~\citep{liu2021swin} leads the integration of hierarchical structure into ViT models, introducing shifted window attention and subsampling techniques. 
PVT~\citep{wang2021pyramid} and PVTv2~\citep{Wang_2022} address these issues using a pyramid structure, linear complexity attention, and overlapping patch embeddings. 
FDViT~\citep{inproceedings}  introduces an innovative downsampling module to address the issue of excessive information loss caused by traditional integer-stride downsampling, and allows the reduction of feature map dimensions with non-integer strides, thereby smoothly decreasing computational complexity while preserving more information.
The Twins~\citep{chu2021twins}  architecture introduces two hierarchical ViT variants: Twins-PCPVT and Twins-SVT. Twins-PCPVT improves the performance of PVT~\citep{wang2021pyramid} through conditional positional encoding, and Twins-SVT employs a robust local-global decoupled SA mechanism, enabling exceptional local detail capture and global information processing. 
MaxViT~\citep{tu2022maxvit} introduces multi-axis attention blocks (Max-SA) to blend local and global paradigms, maintaining efficacy at different resolutions. 
H-MHSA~\citep{liu2024vision} further refines these approaches by maintaining granularity while modeling global and local dependencies. CvT\footnote{\url{https://github.com/microsoft/CvT}}~\citep{Wu_2021_ICCV} takes a different route by integrating convolutions directly into the vision transformer architecture, using the hierarchical feature extraction capabilities of the CNN models to create a multi-stage structure. This approach bridges the gap between the CNN and ViT models, combining their strengths. Similarly, PiT\footnote{\url{https://github.com/naver-ai/pit}}~\citep{heo2021rethinking} proposes methods to better handle spatial dimensions, improving the hierarchical processing of spatial information.

\noindent 
\textbf{{Other Framework.}}
For the efficiency exploration aspect of the novelty structure, Mask R-CNN~\citep{li2021benchmarking} demonstrates how strategic partitioning can enhance task performance through various upsampling or downsampling operations. ViT-Adapter~\citep{chen2023vision} introduces a pre-training-free adapter for more effective inductive bias transfer. Using self-supervised techniques, MixMAE~\citep{liu2023mixmae} by using visible tokens from different images for reconstruction reduces computational demand. 
HiViT~\citep{zhang2022hivit} improves efficiency by excluding all masked tokens. Complementary approaches, such as the asymmetric encoder-decoder structure~\citep{huang2022green}, significantly reduce memory usage by clustering attention and sparse convolutions. Hiera~\citep{ryali2023hiera} advances these ideas by removing non-transformer elements and enhancing image recognition analysis.

\subsubsection{\textbf{Summary: Framework Design}}
Traditional ViTs generate single-resolution feature maps, limiting multi-scale information capture for image recognition. 
By introducing a hierarchical structure design into the ViT architecture, it gains the ability to extract fine-grained features and effectively overcomes isotropy, enabling it to adapt to a diverse range of recognition tasks.
We provide relevant experimental results in Table~\ref{table:hierarchical structures}, where it can be observed that models utilizing hierarchical structures for multi-scale feature extraction generally exhibit superior performance in terms of accuracy. 
In particular, the SOTA Hiera-H~\citep{ryali2023hiera} not only possesses exceptional parameters and FLOPs, but also achieves the highest accuracy of $86.9\%$. 
HAT-Net and Hiera offer diverse optimizations in parameters and computational complexity compared to traditional ViTs and can be flexibly adapted to models of varying capacities to accommodate practical application requirements.

\subsection{\textbf{Efficient ViT Component Discussion}}
In this section, we focus on the internal components of the vision transformer and the efforts made to achieve a balance between efficiency and precision. We present the methods in the order of the image input flow through the network, specifically highlighting four aspects of technical innovation: Embedding Structure Design, Efficient Position Encoding, Efficient Token Update, and Framework Design. In addition, we discuss the relevance of these components in the broader scope of image recognition, emphasizing how they contribute to the overall accuracy and speed of the model.

Tables~\ref{table:patch embedding},~\ref{table:self attention},~\ref{table:MLP} and~\ref{table:hierarchical structures} present the precision benchmarks of these methods and its variants on the ImageNet dataset, highlighting the trade-offs achieved between performance and efficiency at a consistent resolution of $224 \times 224$.  
From these tables, several conclusions can be drawn:
\begin{enumerate}[label=\alph*.]
    \item Advancements in patch embedding and positional encoding, often implemented independently, flexibly integrate into existing frameworks, mitigating the need for extensive hyperparameter tuning and specialized optimization techniques;
    \item Locality emerges as a driving force in enhancing ViT component performance. This is evident in patch embedding that focuses on local regions through the attention mechanism, the reduction of self-attention scope or integration of CNN models, and the hierarchical information extraction in framework design. 
    These strategies yield \ significant efficiency gains without compromising performance;
    \item While shifting from self-attention to attention-free designs demonstrably reduces FLOPs, recent self-attention optimization designs demonstrate a compelling balance between performance and efficiency.
    Therefore, exploring optimization strategies targeting the complexity of self-attention remains a promising direction for future research;
    \item The performance gains observed in hierarchical structures for dense prediction tasks are particularly noteworthy when compared to recognition tasks, suggesting that developing a general hierarchical architecture for multiple vision tasks is a research direction worthy of further exploration. 
\end{enumerate}

In this section, we follow the sequence in which the input stream is forwarded within  the ViT models to present the methods involved, offering the advantage of:
\begin{enumerate}[label=\alph*.]
    \item By analyzing the performance bottlenecks of existing models, we can determine which component corresponds to lightweight measures. This section facilitates this process by providing a straightforward and accessible comparative framework.
    \item By emphasizing the relative independence of ViT components, this offers a perspective on adaptive and integrated optimization, which allows the selection of multiple components along the input stream to achieve lightweight design.
\end{enumerate}

\begin{figure*}[tb]
    \centering
    \includegraphics[width=\textwidth]{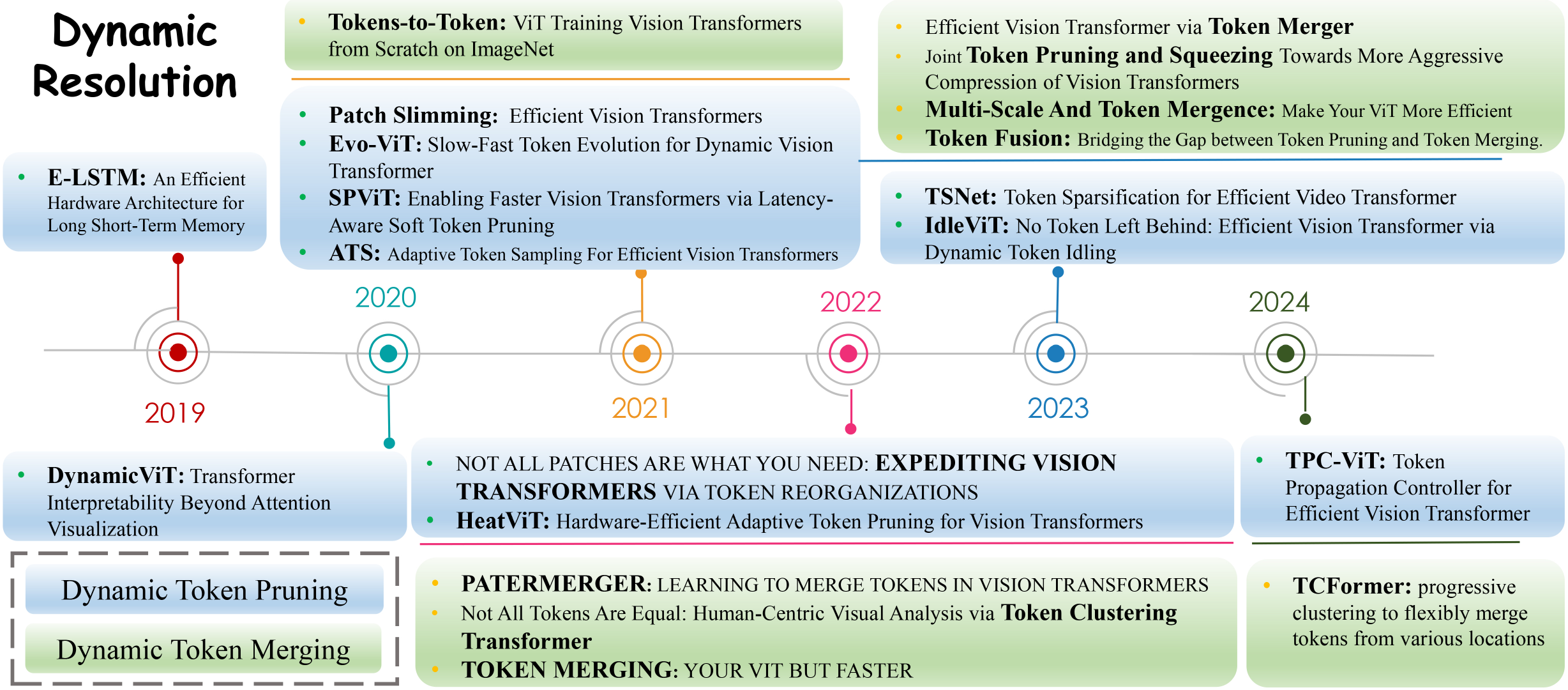}
    \caption{
    \textbf{Dynamic Resolution Exploration Timeline.}
    The figure collects and organizes the development history of dynamic resolution methods in ViT model lightweighting. It is mainly divided into two parts: token pruning and token merging.
    }
    \label{fig:timeline_token}
\end{figure*}

\begin{figure}[tb]
    \centering
    \includegraphics[width=0.95\linewidth]{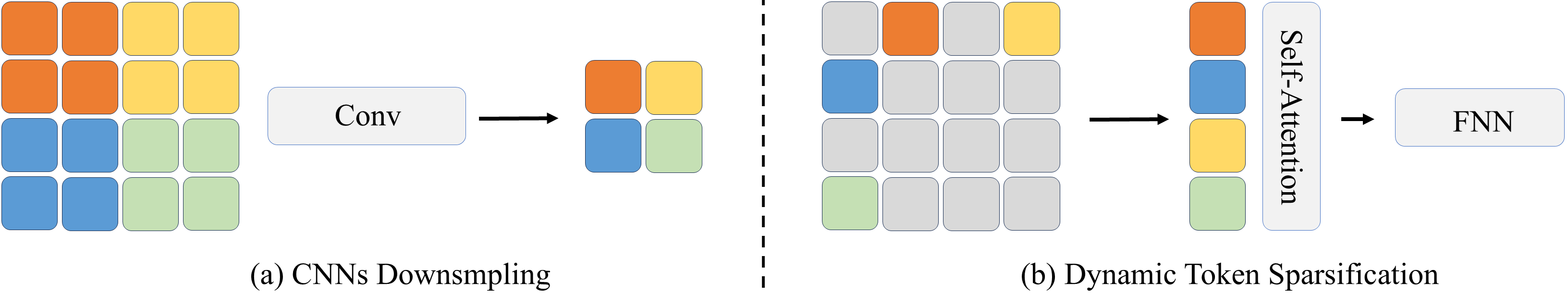}
    \caption{
    \textbf{Space Redundancy Solution Comparison.}
    The CNN models typically employ a structural downsampling strategy, as illustrated in $(a)$. The unstructured and data-dependent downsampling method in $(b)$ more effectively capitalizes on the sparsity within the input data. Owing to the inherent characteristics of the self-attention operation, the unstructured token set is also accelerated through parallel computing.
}
    \label{fig:pruning_merge}
\end{figure}

\begin{figure}[tb]
    \centering
    \includegraphics[width=0.7\linewidth]{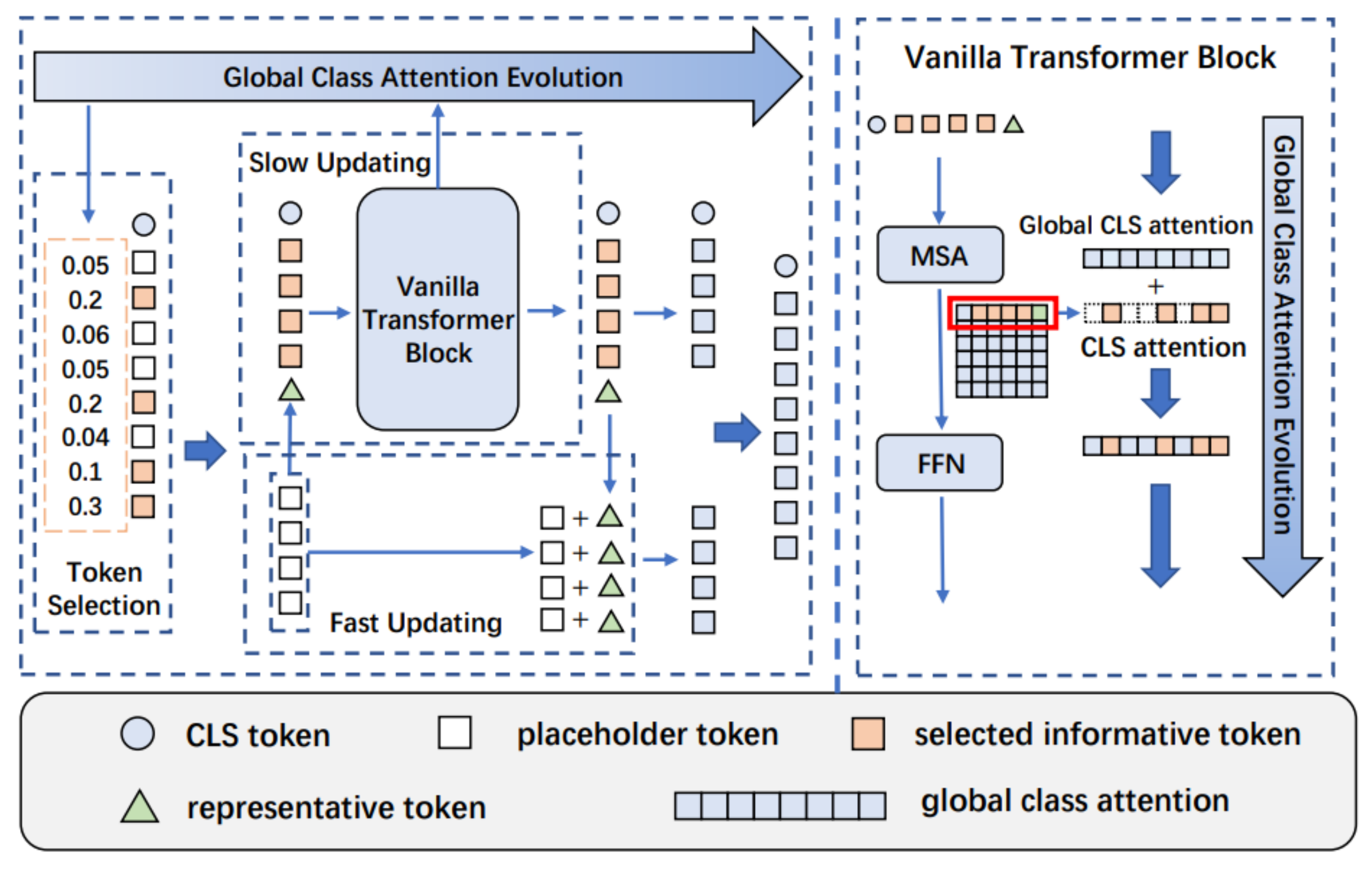}
    \caption{
    \textbf{Token Selection Example}.
    Evo-ViT~\citep{xu2021evovit} uses a token selector to pick informative tokens for slow updates and less-informative tokens for fast updates.}
    \label{fig:EVO-VIT}
\end{figure}

\section{\textbf{Dynamic Network}}
\label{Dynamic Networks}
Dynamic networks represent a critical evolution in the field of image recognition, dramatically improving precision by flexibly adapting their computational structures to the complexities of varying inputs. Unlike static models, which operate on fixed computational graphs and parameters, dynamic networks can tailor their internal configurations, thus achieving better computational efficiency and adaptability. This adaptability is particularly beneficial in image recognition tasks, where the complexity of images can vary widely. 
Furthermore, the dynamic network paradigm is one of the specific forms of online lightweighting strategies, where the computational load is reduced dynamically to maintain efficiency without compromising accuracy. In this section, we categorize dynamic networks based on the vision transformer into two primary topics to review the rapid advances: \textbf{Dynamic Resolution} in Section~\ref{Dynamic Resolution} and \textbf{Depth Adaptation} in Section~\ref{Depth Adaptation}.

\subsection{\textbf{Dynamic Resolution}}
\label{Dynamic Resolution}
As shown in Figure~\ref{fig:pruning_merge}, unlike traditional convolutional neural networks, which often employ coarse sampling methods when faced with spatial redundancy or the need for downsampling, ViT models possess the unique token representation advantage, allowing for more customized and fine-grained   downsampling strategies. 
Specifically, decisions are made about whether each token is pruned, kept, or merged based on its significance. This approach has proven particularly beneficial for image recognition tasks, where the ability to dynamically adjust the resolution and focus on significant regions of an image can greatly enhance the accuracy and efficiency of the model. From these applications, we summarize two main topics: \textbf{Dynamic Token Pruning} and \textbf{Dynamic Token Merging}.
In Figure~\ref{fig:timeline_token}, we summarize the timeline of significant explorations on these two topics.

\begin{figure*}[!t]
    \centering
    \includegraphics[width=\textwidth]{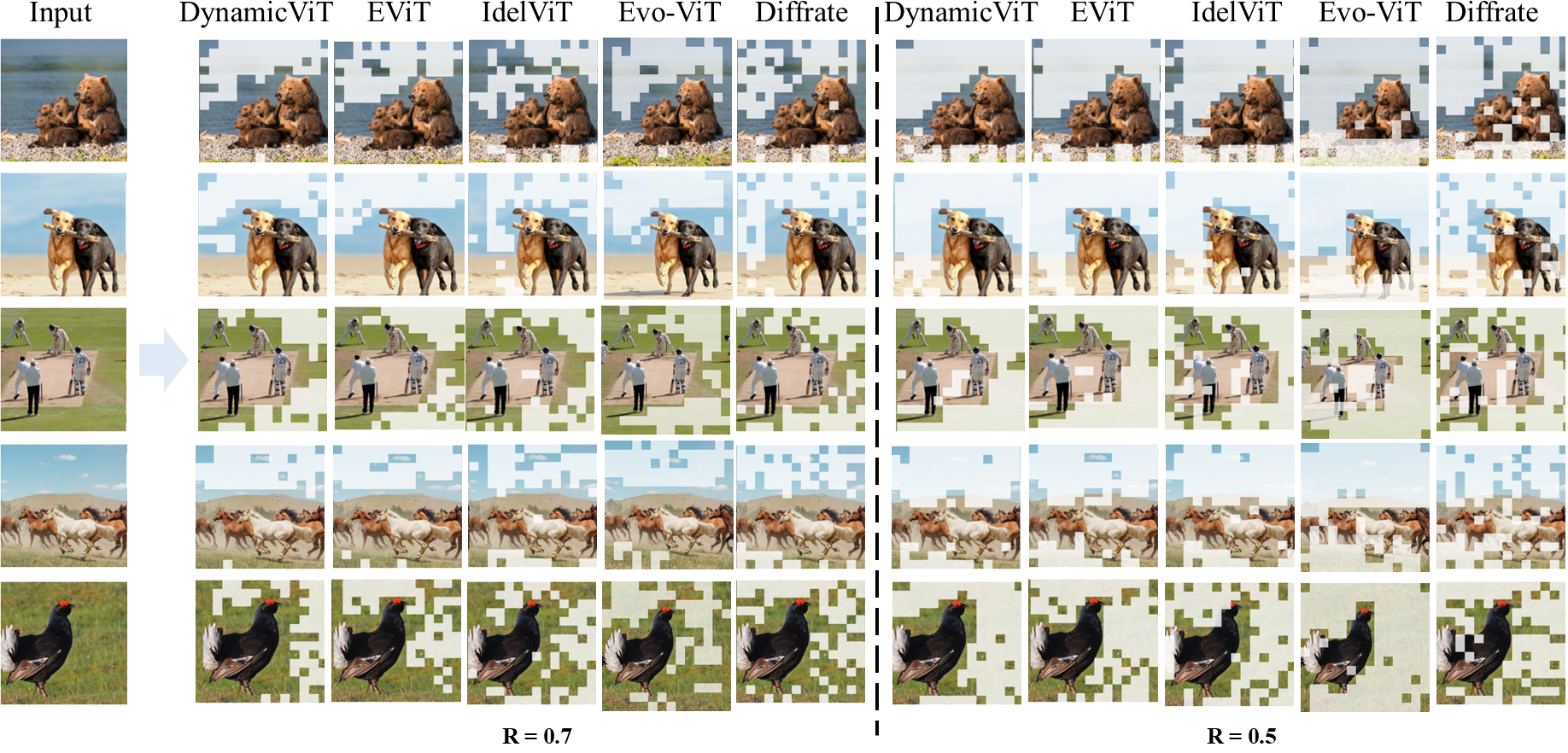}
    \caption{
    \textbf{Token Pruning Visualization}.
    Visualize and compare different token pruning methods. Here, $R$ represents the token keep rate, and the masked tokens are shown on a white background.  
Generally, the higher the attention to the object, the more effective the token pruning method is considered.
    }
    \label{fig:vis_token_pruning}
\end{figure*}

\begin{table*}[tb]
\caption{
A quantitative comparative analysis is conducted on the classic methods involved in \textbf{Dynamic Token Pruning}~(Section~\ref{Dynamic Token Pruning}), using ImageNet-1K as the benchmark, with the resolution standardized to $224 \times 224$.
The R in ``/R" indicates the ratio of token pruning.
The best-performing method is highlighted in bold, and the runner-up is underlined.
}
    \centering
    \begin{minipage}[t]{0.49\linewidth}
    \renewcommand{\arraystretch}{1.34} 
    \resizebox{\linewidth}{!}{
        \begin{tabular}{lcrrrr}
        \bottomrule[1.5px]
         \textbf{Model} &\textbf{Year} & \textbf{\makecell[r]{Params \\ (M)$\downarrow$}} & \textbf{\makecell[r]{Throughput \\ (img/s)$\uparrow$}} & \textbf{\makecell[r]{FLOPs \\ (G)$\downarrow$}}  & \textbf{\makecell[r]{Top-1 Acc. \\ (\%)$\uparrow$}}\\
    \hline
    \tikz\fill[ceiling] (0,0) circle (3.5pt);\:Evo-ViT(DeiT-T)~\cite{xu2021evovit} &2021& 5.9 & 4,027& \thirdcolor{0.8}&72.0\\
    \tikz\fill[ceiling] (0,0) circle (3.5pt);\:Evo-ViT(DeiT-S)~\cite{xu2021evovit} &2021& 22.4 & 1,510& \thirdcolor{3.0}&79.4\\
    \tikz\fill[ceiling] (0,0) circle (3.5pt);\:Evo-ViT(DeiT-B)~\cite{xu2021evovit} &2021& 87.3 & 462& \thirdcolor{11.6}&81.3\\
    \hline
    \rowcolor{gray!20} \tikz\fill[objects] (0,0) circle (3.5pt);DynamicViT(DeiT-T)/0.7~\cite{zheng2022dynamic} &2021& 5.9 & \thirdcolor{2,581} & \thirdcolor{1.0} &71.0\\
    \rowcolor{gray!20} \tikz\fill[objects] (0,0) circle (3.5pt);\:DynamicViT(DeiT-T)/0.5~\cite{zheng2022dynamic} &2021& 5.9 & \thirdcolor{2,590} & \thirdcolor{0.8} & 67.4\\
    \rowcolor{gray!20} \tikz\fill[objects] (0,0) circle (3.5pt);\:DynamicViT(DeiT-S)/0.9~\cite{zheng2022dynamic} &2021& \thirdcolor{22.8} & 1,524 & 4.0 & 79.8\\
    \rowcolor{gray!20} \tikz\fill[objects] (0,0) circle (3.5pt);\:DynamicViT(DeiT-S)/0.7~\cite{zheng2022dynamic} &2021& \thirdcolor{22.8} & 2,062 & 2.9 &79.3\\
    \rowcolor{gray!20} \tikz\fill[objects] (0,0) circle (3.5pt);\:DynamicViT(DeiT-T)/0.7~\cite{zheng2022dynamic}  &2021& \thirdcolor{89.4} & \thirdcolor{373} & \thirdcolor{11.2} &81.3\\
    \hline
    \tikz\fill[sofa] (0,0) -- (8pt,0) -- (4pt,8pt) -- cycle;\:EViT-Tiny~\cite{liang2022patches} &2022& 12.1 & \thirdcolor{3,568} & 1.9 &79.9\\
    \tikz\fill[sofa] (0,0) -- (8pt,0) -- (4pt,8pt) -- cycle;\:EViT-Small~\cite{liang2022patches} &2022& 23.4 & \thirdcolor{2,007} & 3.4 &82.6\\
    \tikz\fill[sofa] (0,0) -- (8pt,0) -- (4pt,8pt) -- cycle;\:EViT-Base~\cite{liang2022patches} &2022& 42.6 & \thirdcolor{1,008} & 6.3 &\underline{83.9}\\
     \tikz\fill[sofa] (0,0) -- (8pt,0) -- (4pt,8pt) -- cycle;\:EViT-Large~\cite{liang2022patches} &2022& 60.1 & - & 9.4 &\textbf{84.4}\\
    \hline
    \rowcolor{gray!20} \tikz\fill[gray] (0,0) -- (8pt,0) -- (4pt,8pt) -- cycle;\:SPViT(DeiT-T)/1.0~\cite{kong2022spvit} &2022& 5.7 & \thirdcolor{1,372} & 1.3 &72.2\\
    \rowcolor{gray!20} \tikz\fill[gray] (0,0) -- (8pt,0) -- (4pt,8pt) -- cycle;\:SPViT(DeiT-T)/0.7~\cite{kong2022spvit} &2022& 5.7 & \thirdcolor{1,680} & \thirdcolor{0.8} &72.1\\
    \rowcolor{gray!20} \tikz\fill[gray] (0,0) -- (8pt,0) -- (4pt,8pt) -- cycle;\:SPViT(DeiT-S)/1.0~\cite{kong2022spvit} &2022& 22.0 & \thirdcolor{718} & \thirdcolor{4.6} &79.9\\
    \rowcolor{gray!20} \tikz\fill[gray] (0,0) -- (8pt,0) -- (4pt,8pt) -- cycle;\:SPViT(DeiT-S)/0.7~\cite{kong2022spvit} &2022& 22.0 & \thirdcolor{947} & \thirdcolor{2.9} &79.5\\
    \rowcolor{gray!20} \tikz\fill[gray] (0,0) -- (8pt,0) -- (4pt,8pt) -- cycle;\:SPViT(DeiT-B)/1.0~\cite{kong2022spvit} &2022& 86.6 & \thirdcolor{246} & \thirdcolor{17.7} &82.2\\
    \rowcolor{gray!20} \tikz\fill[gray] (0,0) -- (8pt,0) -- (4pt,8pt) -- cycle;\:SPViT(DeiT-B)/0.7~\cite{kong2022spvit} &2022& 86.6 & \thirdcolor{322} & \thirdcolor{11.1} &82.0\\
    \bottomrule[1.5px]
    \end{tabular}
        }
    \end{minipage}
    \hfill
    \centering
     \begin{minipage}[[t]{0.49\linewidth}
     \renewcommand{\arraystretch}{1.35} 
     \resizebox{\linewidth}{!}{
        \begin{tabular}{lcrrrr}
        \bottomrule[1.5px]
         \textbf{Model} &\textbf{Year} & \textbf{\makecell[r]{Params \\ (M)$\downarrow$}} & \textbf{\makecell[r]{Throughput \\ (img/s)$\uparrow$}} & \textbf{\makecell[r]{FLOPs \\ (G)$\downarrow$}}  & \textbf{\makecell[r]{Top-1 Acc. \\ (\%)$\uparrow$}}\\
    \hline
    \tikz\fill[trunk] (0,0) rectangle (8pt,6pt);\:HeatViT-T0~\cite{10071047} &2023& - & \thirdcolor{4,172} & 0.5 &70.8\\
    \tikz\fill[trunk] (0,0) rectangle (8pt,6pt);\:HeatViT-T2~\cite{10071047} &2023& - & \thirdcolor{3,200} & 1.0 &72.2\\
     \tikz\fill[trunk] (0,0) rectangle (8pt,6pt);\:HeatViT-S0~\cite{10071047} &2023& - & \thirdcolor{1,390} & 1.8 &78.5\\
    \tikz\fill[trunk] (0,0) rectangle (8pt,6pt);\:HeatViT-S1~\cite{10071047} &2023& - & \thirdcolor{1,860} & 2.0 &79.0\\
    \tikz\fill[trunk] (0,0) rectangle (8pt,6pt);\:HeatViT-B0~\cite{10071047} &2023& - & \thirdcolor{548} & 6.1 &80.1\\
    \tikz\fill[trunk] (0,0) rectangle (8pt,6pt);\:HeatViT-B1~\cite{10071047} &2023& - & \thirdcolor{415} & 10.5 &81.1\\
    \hline
    \rowcolor{gray!20} \tikz\fill[parking] (0,0) rectangle (8pt,6pt);\:CF-ViT(DeiT-S)/1.0~\cite{chen2022cfvit} &2023& 22.1 & 2,760 & 4.0 &80.8\\
    \rowcolor{gray!20} \tikz\fill[parking] (0,0) rectangle (8pt,6pt);\:CF-ViT(DeiT-S)/0.75~\cite{chen2022cfvit} &2023& 22.1 & 3,701 & 2.6 &80.7\\
    \rowcolor{gray!20} \tikz\fill[parking] (0,0) rectangle (8pt,6pt);\:CF-ViT(DeiT-S)/0.5~\cite{chen2022cfvit} &2023& 22.1 & 4,903 & 1.8 &79.8\\
    \hline
    \tikz\fill[other_veh] (0,0) rectangle (8pt,6pt);\:IdleViT(DeiT-S)/0.9~\cite{xu2023token} &2023& 22.1 &2,662&4.0&79.9\\
    \tikz\fill[other_veh] (0,0) rectangle (8pt,6pt);\:IdleViT(DeiT-S)/0.7~\cite{xu2023token} &2023& 22.1 &3,361&3.1&79.6\\
    \tikz\fill[other_veh] (0,0) rectangle (8pt,6pt);\:IdleViT(DeiT-S)/0.5~\cite{xu2023token} &2023& 22.1 &4,071&2.4&79.0\\
   
    \hline
    \rowcolor{gray!20} \tikz\fill[tvs] (0,0) -- (8pt,0) -- (4pt,8pt) -- cycle;\:ATS(DeiT-S)~\cite{fayyaz2022adaptive}&2022& 22.0 & 1,403 & \thirdcolor{2.9} &79.7\\
   \rowcolor{gray!20} \tikz\fill[tvs] (0,0) -- (8pt,0) -- (4pt,8pt) -- cycle;\:ATS(CvT-13)~\cite{fayyaz2022adaptive}&2022& 22.0 & \thirdcolor{1,080} & \thirdcolor{3.2} &81.4\\
    \hline
    \tikz\fill[truck] (0,0) rectangle  (6.5pt,6.5pt);\:TPC-ViT(DeiT-T)~\cite{zhu2024tpcvit} &2024& 5.7 & \thirdcolor{4,500} & 0.6&73.0\\
    \tikz\fill[truck] (0,0) rectangle  (6.5pt,6.5pt);\:TPC-ViT(DeiT-S)~\cite{zhu2024tpcvit} &2024& 22.0 & \thirdcolor{2,303} & 2.8&80.8\\
    \tikz\fill[truck] (0,0) rectangle  (6.5pt,6.5pt);\:TPC-ViT(DeiT-B)~\cite{zhu2024tpcvit} &2024& 86.6 & \thirdcolor{490} & 10.7&81.8\\
    \bottomrule[1.5px]
        \end{tabular}
        }
    \end{minipage}
    \label{table:token pruning}
\end{table*}

\begin{table*}[tb]
\caption{
A quantitative comparative analysis is conducted on the classic methods involved in \textbf{Dynamic Token Merging}~(Section~\ref{Dynamic Token Merging}), using ImageNet-1K as the benchmark, with the resolution standardized to $224 \times 224$.
The R in ``/R” shows the token merging rate. 
If $R<1$, the model keeps R\% of the original tokens. If $R>1$, the model is tested after merging tokens into R blocks.
The best-performing method is highlighted in bold, and the runner-up is underlined.
}
    \centering
    \begin{minipage}[t]{0.49\linewidth}
    \renewcommand{\arraystretch}{1.34} 
    \resizebox{\linewidth}{!}{
        \begin{tabular}{lcrrrr}
        \bottomrule[1.5px]
         \textbf{Model} &\textbf{Year} & \textbf{\makecell[r]{Params \\ (M)$\downarrow$}} & \textbf{\makecell[r]{Throughput \\ (img/s)$\uparrow$}} & \textbf{\makecell[r]{FLOPs \\ (G)$\downarrow$}}  & \textbf{\makecell[r]{Top-1 Acc. \\ (\%)$\uparrow$}}\\
    \hline

    \tikz\fill[ceiling] (0,0) circle (3.5pt);\:T2T-ViT(ViT-14)~\cite{Yuan_2021_ICCV} & 2021 & 21.5 & 1,944 & 4.8 & 81.5 \\
    \tikz\fill[ceiling] (0,0) circle (3.5pt);\:T2T-ViT(ViT-19)~\cite{Yuan_2021_ICCV} & 2021 & 39.2 & 1,660 & 8.5 & 81.9 \\
    \tikz\fill[ceiling] (0,0) circle (3.5pt);\:T2T-ViT(ViT-24)~\cite{Yuan_2021_ICCV} & 2021 & 64.1 & 1,497 & 13.8 & 82.3 \\
    \hline
    \rowcolor{gray!20} \tikz\fill[parking] (0,0) rectangle (8pt,6pt);\:MHSA(DeiT-T)~\cite{bian2023multiscale} & 2023 & 5.8 &- & 0.8 &72.7\\
    \rowcolor{gray!20} \tikz\fill[parking] (0,0) rectangle (8pt,6pt);\:MHSA(DeiT-S)~\cite{bian2023multiscale} & 2023 & 22.2 & - & 3.1 &79.7\\
    \rowcolor{gray!20} \tikz\fill[parking] (0,0) rectangle (8pt,6pt);\:MHSA(LV-ViT-S)~\cite{bian2023multiscale} & 2023 & 26.2 & - & 3.9 &82.8\\
    \rowcolor{gray!20} \tikz\fill[parking] (0,0) rectangle (8pt,6pt);\:MHSA(DeiT-B)~\cite{bian2023multiscale} & 2023 & 86.6 & - & 11.5 &81.5 \\
    \hline
    \tikz\fill[trunk] (0,0) rectangle (8pt,6pt);\:SuperViT(LV-ViT-T)/1.0~\cite{lin2023super} & 2023 & \thirdcolor{5.5} & 3,178 & 2.9 &79.9 \\
    \tikz\fill[trunk] (0,0) rectangle (8pt,6pt);\:SuperViT(DeiT-S)/0.5~\cite{lin2023super} & 2023 & \thirdcolor{5.7} & 4,767 & 2.3 &80.0\\
    \tikz\fill[trunk] (0,0) rectangle (8pt,6pt);\:SuperViT(DeiT-T)/1.0~\cite{lin2023super} & 2023 & \thirdcolor{5.7} & 5,013& 1.3 &73.1\\
    \tikz\fill[trunk] (0,0) rectangle (8pt,6pt);\:SuperViT(DeiT-S)/0.7~\cite{lin2023super} & 2023 & \thirdcolor{22.3} & 3,654 & 3.0 &80.5\\
     \tikz\fill[trunk] (0,0) rectangle (8pt,6pt);\:SuperViT(DeiT-S)/1.0~\cite{lin2023super} & 2023 & \thirdcolor{22.3} & 2,461 & 4.6 &80.6\\
     \tikz\fill[trunk] (0,0) rectangle (8pt,6pt);\:SuperViT(LV-ViT-S)/0.7~\cite{lin2023super} & 2023 & \thirdcolor{27.8} & 2,684 & 4.3 &83.2\\
    \tikz\fill[trunk] (0,0) rectangle (8pt,6pt);\:SuperViT(LV-ViT-S)/1.0~\cite{lin2023super} & 2023 & \thirdcolor{27.8} & 1,748 & 6.6 &83.5\\
    \hline
    \rowcolor{gray!20} \tikz\fill[other_veh] (0,0) rectangle (8pt,6pt);\:ToMe(ViT-B)/8~\cite{bolya2023token} & 2023 & 86.0 & 413 & 13.1 & 82.9\\
    \bottomrule[1.5px]
    \end{tabular}
        }
    \end{minipage}
    \hfill
    \centering
     \begin{minipage}[[t]{0.49\linewidth}
     \renewcommand{\arraystretch}{1.39} 
     \resizebox{\linewidth}{!}{
        \begin{tabular}{lcrrrr}
        \bottomrule[1.5px]
         \textbf{Model} &\textbf{Year} & \textbf{\makecell[r]{Params \\ (M)$\downarrow$}} & \textbf{\makecell[r]{Throughput \\ (img/s)$\uparrow$}} & \textbf{\makecell[r]{FLOPs \\ (G)$\downarrow$}}  & \textbf{\makecell[r]{Top-1 Acc. \\ (\%)$\uparrow$}}\\
    \hline
    
    \rowcolor{gray!20} \tikz\fill[other_veh] (0,0) rectangle (8pt,6pt);\:ToMe(ViT-B)/12~\cite{bolya2023token} & 2023 & 86.0 & 489 & 10.9 & 81.8\\
    \rowcolor{gray!20} \tikz\fill[other_veh] (0,0) rectangle (8pt,6pt);\:ToMe(ViT-B)/16~\cite{bolya2023token} & 2023 & 86.0 & 607 & 8.7 & 78.9\\
    \rowcolor{gray!20} \tikz\fill[other_veh] (0,0) rectangle (8pt,6pt);\:ToMe(ViT-B)/20~\cite{bolya2023token} & 2023 & 86.0 & 736 & 7.1 & 67.5\\
    \rowcolor{gray!20} \tikz\fill[other_veh] (0,0) rectangle (8pt,6pt);\:ToMe(ViT-L)/8~\cite{bolya2023token} & 2023 & 307.0 & 182 & 61.6 & \textbf{84.2}\\
    \hline
    \tikz\fill[truck] (0,0) rectangle  (6.5pt,6.5pt);\:ToFu AVG(ViT-B)/8~\cite{kim2023token} & 2023 & \thirdcolor{86.6} & 417 & 13.1 & 83.2\\
    \tikz\fill[truck] (0,0) rectangle  (6.5pt,6.5pt);\:ToFu AVG(ViT-B)/12~\cite{kim2023token} & 2023 & \thirdcolor{86.6} & 484 & 10.9 & 82.4\\
    \tikz\fill[truck] (0,0) rectangle  (6.5pt,6.5pt);\:ToFu AVG(ViT-B)/16~\cite{kim2023token} & 2023 & \thirdcolor{86.6} & 615 & 8.7 & 80.4\\
    \tikz\fill[truck] (0,0) rectangle  (6.5pt,6.5pt);\:ToFu AVG(ViT-B)/20~\cite{kim2023token} & 2023 & \thirdcolor{86.6} & 748 & 7.1 & 72.2\\
    \hline
    \rowcolor{gray!20} \tikz\fill[sofa] (0,0) -- (8pt,0) -- (4pt,8pt) -- cycle;\:TCFormerV1-Light~\cite{zeng2022tokens} & 2024 & 14.2 & 185 & 3.8 &79.6\\
    \rowcolor{gray!20} \tikz\fill[sofa] (0,0) -- (8pt,0) -- (4pt,8pt) -- cycle;\:TCFormerV2-Tiny~\cite{zeng2022tokens} & 2024 & 14.2 & 393 & 2.5 &79.5\\
    \rowcolor{gray!20} \tikz\fill[sofa] (0,0) -- (8pt,0) -- (4pt,8pt) -- cycle;\:TCFormerV1~\cite{zeng2022tokens} & 2024 & 25.6 & 120 & 5.9 &82.4\\
    \rowcolor{gray!20} \tikz\fill[sofa] (0,0) -- (8pt,0) -- (4pt,8pt) -- cycle;\:TCFormerV2-Small~\cite{zeng2022tokens} & 2024 & 25.6 & 273 & 4.5 &82.4\\
    \rowcolor{gray!20} \tikz\fill[sofa] (0,0) -- (8pt,0) -- (4pt,8pt) -- cycle;\:TCFormerV1-Large~\cite{zeng2022tokens} & 2024 & 62.8 & 58 & 12.2 &83.6\\
    \rowcolor{gray!20}  \tikz\fill[sofa] (0,0) -- (8pt,0) -- (4pt,8pt) -- cycle;\:TCFormerV2-Base~\cite{zeng2022tokens} & 2024 & 62.8 & 103 & 10.8 & \underline{83.8} \\
    \bottomrule[1.5px]
        \end{tabular}
        }
    \end{minipage}
    \label{table:token merge}
\end{table*}

\subsubsection{\textbf{Dynamic Token Pruning}}
\label{Dynamic Token Pruning}
In relation to this topic, tokens with insufficient informational content are often directly pruned or sent through a different update path, whereas tokens recognized as having rich semantic information are passed to the deeper layers of the network.

The Top-K pruning method \citep{8693509} keeps only the $K$ most significant tokens based on their attention weights. 
PS-ViT\footnote{\url{https://github.com/yuexy/PS-ViT}}~\citep{chen2021psvitbettervisiontransformer} reduces tokens according to their measured importance to minimize computational costs while retaining performance. 
Patch thinning \citep{tang2022patch} employs a top-down strategy, with later layers guiding the token selection process in earlier layers.
EViT~\footnote{\url{https://github.com/youweiliang/evit}} \citep{liang2022patches} fuses less important tokens into a single token for further computation, preserving the essence while reducing the number.
Evo-ViT \citep{xu2021evovit} is shown in Figure~\ref{fig:EVO-VIT}, regulates the flow of gradient propagation, updating critical tokens slowly and less important ones quickly before merging. IdleViT \citep{xu2023token} implements a token idling strategy, processing only a subset of tokens in each layer and passing the rest to the next layer unchanged.

Recent studies emphasize the importance of token selection mechanisms adapted to evolving ViT architectures. SPViT \citep{kong2022spvit} employs a dynamic multi-head token selector to filter and integrate less informative tokens. 
The token compensator (ToCom) \citep{Jie2024TokenCA} proposes a model arithmetic framework to decouple compression degrees between the training and inference stages, addressing performance drops caused by mismatched compression degrees, and achieving universal performance improvements without further training. 
HeatViT \citep{10071047} introduces a delay-aware, multi-stage training strategy to optimize token selector performance. TSNet \citep{app131910633} dynamically selects informative tokens from video samples using a lightweight scoring module, while DynamicViT \citep{rao2021dynamicvit} incorporates a learnable module within each transformer block for layer-by-layer token pruning. 
TPC \citep{zhu2024tpcvit} uses suspension and reuse probabilities in token selection to preserve computational accuracy. Lastly, the Adaptive Token Sampler (ATS) \citep{fayyaz2022adaptive} offers a differentiable parameter-free method for adaptive downsampling, minimizing information loss. 
In Figure~\ref{fig:vis_token_pruning}, we present a visual comparison of token pruning strategy methods, generally indicating that the degree of focus on objects demonstrates the ability of the method to identify regions crucial to decision-making.

\begin{figure*}[!t]
    \centering
    \includegraphics[width=\textwidth]{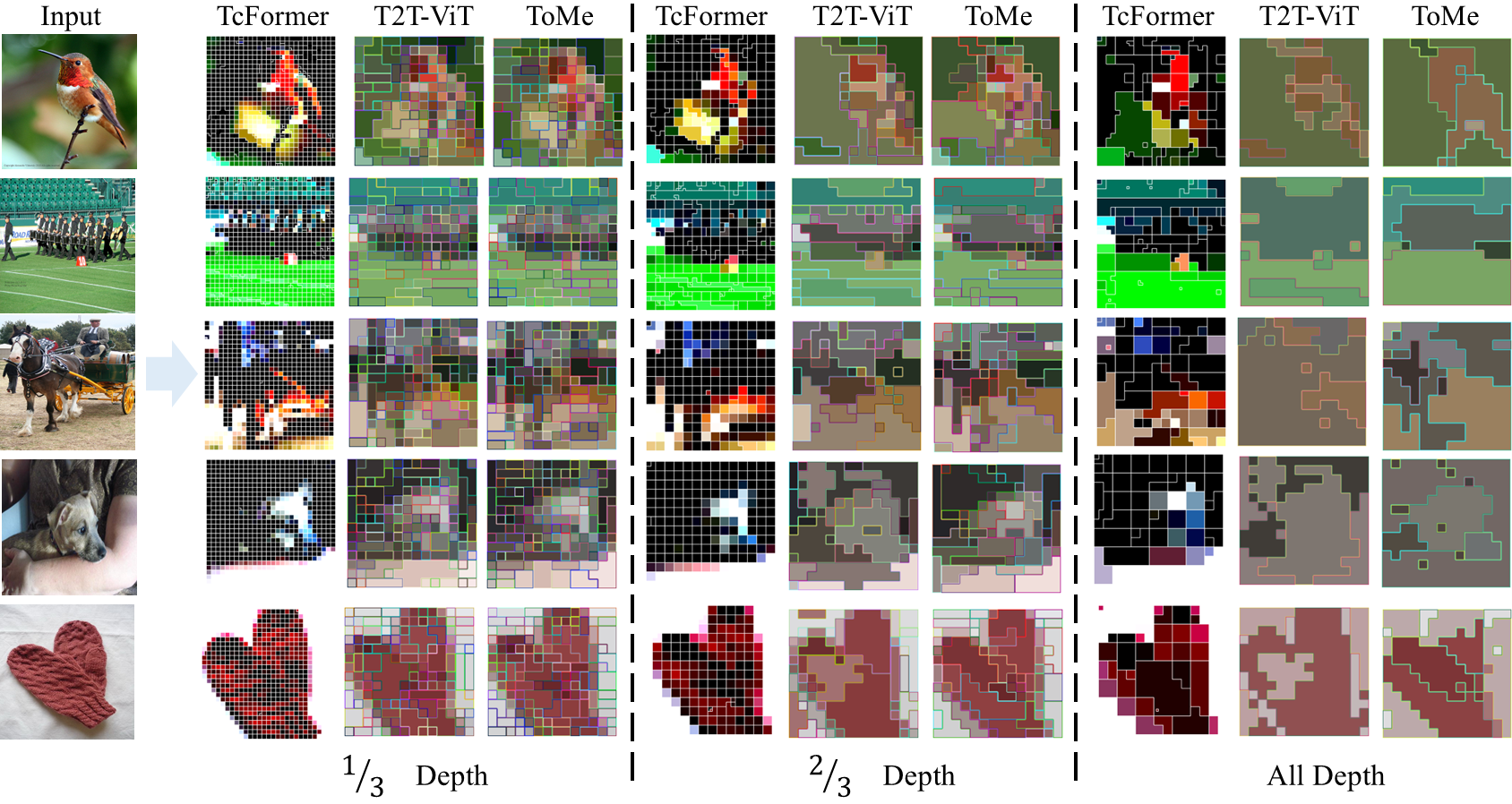}
    \caption{
    \textbf{Token Merge Visualization}.
    Visualize the comparison of token merging methods, where $``1/3"$, $``2/3"$, and ``All" indicates the stage within the ViT stack structure where the token merging operation is executed. Typically, the shallower layers focus on merging detailed texture information, while the deeper layers attend to the merging of more abstract semantic information.
}
    \label{fig:vis_token_merge}
\end{figure*}

\subsubsection{\textbf{Dynamic Token Merging}}
\label{Dynamic Token Merging}
Unlike  pruning methods that discard tokens directly, token merging through strategic token merging, this paradigm can combine less informative tokens with more critical ones, reduce the number of tokens processed, thus contributing to the broader goal of model simplification.

ToMe~\citep{bolya2023token}\footnote{\url{https://github.com/facebookresearch/ToMe} } introduces a fine-tuned matching algorithm to merge similar tokens across layers. 
PatchMerger~\citep{renggli2022learning} allows the network to handle varying token counts by decoupling the number of output tokens from the input, increasing flexibility. \citep{wei2023joint} presents the token pruning and squeezing~(TPS) module, which uses a token merging strategy based on nearest neighbor matching and similarity. 
MHSA~\citep{bian2023multiscale} proposes merging less critical tokens before pruning to maintain crucial information and prevent performance drops. \citep{10183862} devises a meta-token mechanism to generate an attention map to merge similar tokens. T2T-ViT~\citep{DBLP:journals/corr/abs-2101-11986} rearranges the input token sequence to focus on important local structures, facilitating effective token merging. TCFormer~\citep{zeng2022tokens} uses progressive clustering to flexibly merge tokens from various locations, capturing detailed information. Lastly, Token Fusion~\citep{kim2023token} dynamically combines pruning and merging based on token computation.
Figure~\ref{fig:vis_token_merge} presents a visual comparison of token merging strategies, illustrating how merging is performed at varying depths within the ViT architecture.
It is especially noticeable that the initial layers primarily merge fine-grained texture details, whereas the subsequent layers are more focused on integrating higher-level semantic information.

\subsubsection{\textbf{Summary: Dynamic Resolution}}
Eliminating spatial redundancy enables ViT to offer lightweight online design. By evaluating token importance, it retains, prunes, or merges tokens, ideal for real-time image recognition on high-resolution images or large-scale datasets.
Relevant experimental analyzes are provided in Table ~\ref{table:token pruning} and Table ~\ref{table:token merge}, from which it can be observed that:
\textit{\textbf{(i).}} The dynamic resolution strategy can still achieve a lightweight extreme. 
For example, Evo-ViT~\citep{xu2021evovit} has only 5.9M parameters and has a high throughput.
\textit{\textbf{(ii).}} Most methods can significantly improve throughput and reduce FLOPs while slightly decreasing accuracy. For example, DynamicViT~\citep{rao2021dynamicvit} increases throughput by $36.6\%$ and reduces FLOPs by $27.5\%$ compared to the original DeiT-S model, with only the $0.5\%$ accuracy drop.
\textit{\textbf{(iii).}} Larger models (such as ViT-B, ViT-L) can typically tolerate more aggressive token reduction while maintaining high accuracy. For example, ToMe~\citep{bolya2023token} substantially reduces computational cost while maintaining a high accuracy of $84.2\%$.
\textit{\textbf{(iv).}}  Different methods have their own advantages in balancing performance and efficiency. For example, the EViT~\citep{liang2022evit} series excels in maintaining high accuracy, while the SuperViT~\citep{lin2023super} series has more advantages in improving throughput.

\subsection{\textbf{Depth Adaptation}}
\label{Depth Adaptation}
\begin{figure*}[!ht]
    \centering
    \includegraphics[width=0.7\textwidth]{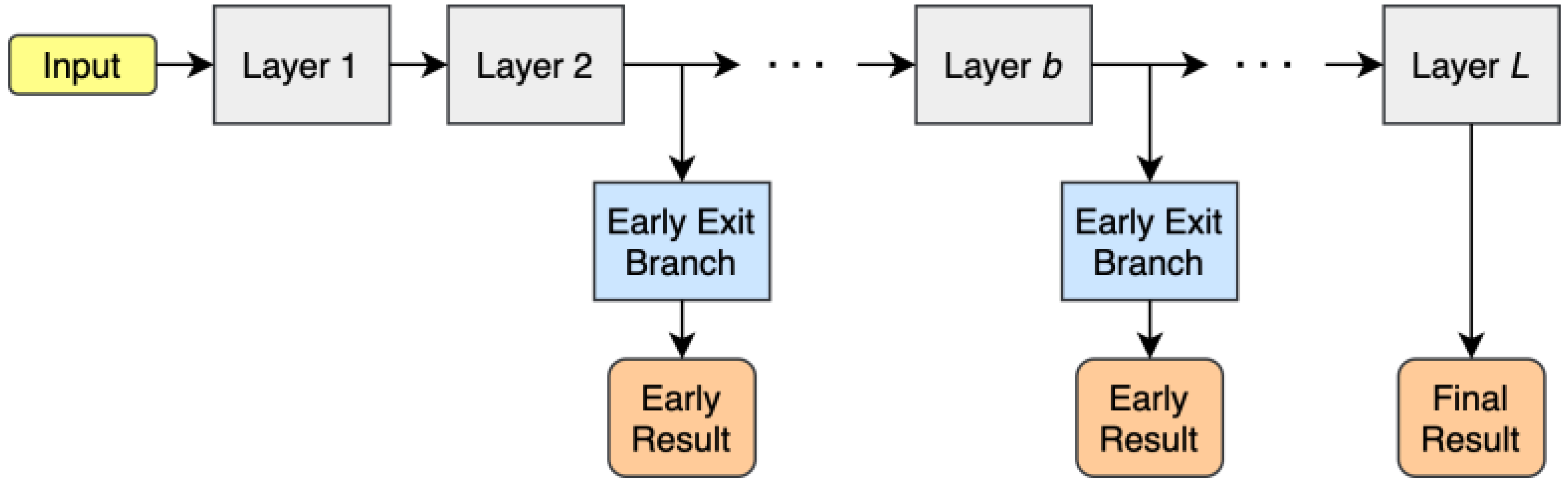}
    \caption{
    \textbf{Early Exit Paradigm.}
    Attaching additional internal classifiers at various stacked layers of the ViT model halts computation upon satisfying predefined exit conditions, thereby facilitating the early exit of simpler samples without compromising image recognition accuracy, which in turn conserves computational resources.
    }
    \label{fig:early exit}
\end{figure*}
Dynamic multi-exit networks provide a flexible online lightweighting technique by enabling early computation halts. 
AViT \citep{Yin_2022_CVPR} adapts ViT to dynamically adjust the depth and computation of the layer based on the specific task and the complexity of the input data.
\citep{9009834} enhances this paradigm using the knowledge distillation training procedure, allowing early exits to benefit from the abstract semantic knowledge of deeper layers.
As shown in Figure~\ref{fig:early exit},\citep{bakhtiarnia2022singlelayer} introduces a novel early exit architecture that uses fine-tuning to significantly improve exit accuracy. \citep{bakhtiarnia2021multiexit} integrates multi-exit architecture into ViT, proposing seven distinct early exit designs. To address performance degradation in early exit methods, LGViT~\citep{Xu_2023} takes advantage of heterogeneous exit headers (local perceptual and global aggregation) to balance efficiency and precision.
DYN-ADAPTER\footnote{\url{https://github.com/LMD0311/DAPT}}~\citep{zhang2024dynadapterdisentangledrepresentationefficient} dynamically adjusts the model capacity based on disentangled representations.
CF-ViT\footnote{\url{https://github.com/ChenMnZ/CF-ViT}}~\citep{chen2022cfvit} divides the processing pipeline into coarse and fine steps. AdaViT\footnote{\url{https://github.com/MengLcool/AdaViT}}~\citep{Meng2021AdaViTAV} adapts the depth adjustment of ViT models based on the complexity of the task.

To further enhance the efficiency and adaptability of the ViT models, several dynamic training strategies are explored. \citep{liu2021faster} presents a faster depth-adaptive transformer, taking advantage of dynamic adjustment of the model layers to optimize computational resources. \citep{elbayad2020depthadaptivetransformer} allows different layers to exit at different points, thus reducing unnecessary computations. \citep{schuster2022confident} proposes a confident adaptive language model that dynamically halts computations based on confidence measures, which can be paralleled in vision tasks to enhance efficiency. Additionally, Skipdecode~\citep{delcorro2023skipdecodeautoregressiveskipdecoding} presents an autoregressive skip decoding mechanism that can potentially be adapted for ViT models to optimize inference efficiency by skipping certain decoding steps while maintaining performance.

\subsubsection{\textbf{Summary: Depth Adaptation}}
In traditional fixed-depth networks, both simple and complex inputs undergo the same computational process, which to some extent wastes resources. 
This topic introduces a dynamic multi-exit paradigm that allows the model to flexibly adjust its computational depth based on input complexity, which aligns with the principles of online lightweighting strategies. 
The multi-exit structure can be regarded as a form of ``implicit ensemble",  where models of different depths share parameters but provide diverse decisions. Moreover, through knowledge distillation techniques, the early layers can also acquire deep semantic information. As ImageNet is a large-scale and inconsistent difficulty dataset, dynamic depth can offer better adaptability, improving overall recognition accuracy. For simple images, results can be obtained at shallower layers, reducing unnecessary computations.

\subsection{\textbf{Dynamic Networks Discussion}}
Based on the complexity of the input and the inherent token representation advantages of ViT, a wide variety of dynamic networks can be constructed. 
Compared to the dynamic network paradigm of the CNN model, ViT has the additional advantage of handling input sequences of variable length. This means ViT does not need to worry about resizing to a rectangular space after eliminating redundancy, thus offering greater flexibility. 
This flexibility is particularly beneficial in lightweight online strategies, where token pruning and merging are employed to reduce computational load while simultaneously enhancing recognition accuracy. 
We report the relevant experimental results in Table~\ref{table:token pruning} and Table~\ref{table:token merge}, and summarize the limitations of the existing methods as follows:
\begin{enumerate}[label=\alph*.]
    \item  Lack of guidance for the pruning or merging layers, with the current widespread practice of manually setting them to be executed at the $3^{th}$, $6^{th}$, and $9^{th}$ layers\citep{liang2022evit,rao2021dynamicvit}. 
    \item Adaptive pruning or merging rates, although learnable pruning rate methods such as DiffRate~\citep{chen2023diffratedifferentiablecompression} are now available, require additional complex optimization techniques and training overhead.
    \item Token pruning or token merging needs to be adapted to subsequent model pruning or quantization, making an end-to-end training strategy necessary.
    \item Although ViT can handle token sequences of variable lengths, token pruning or token merging disrupts position information, necessitating methods for re-modeling position information.
\end{enumerate}
On the other hand, for depth adaptation, the basic paradigm of the early exit strategy in CNN is generally followed, but the following limitations still exist:
\begin{enumerate}[label=\alph*.]
    \item Internal classifiers serve primarily as projections to category space,  failing to fully leverage the advantage of token representations.
    \item Lack of effective guidance for the placement of internal classifiers.
    \item The potential to skip individual components within transformer blocks, such as multi-head self-attention modules or feed-forward networks, rather than entire blocks.  
\end{enumerate}

\begin{figure*}[tb]
    \centering
    \includegraphics[width=\textwidth]{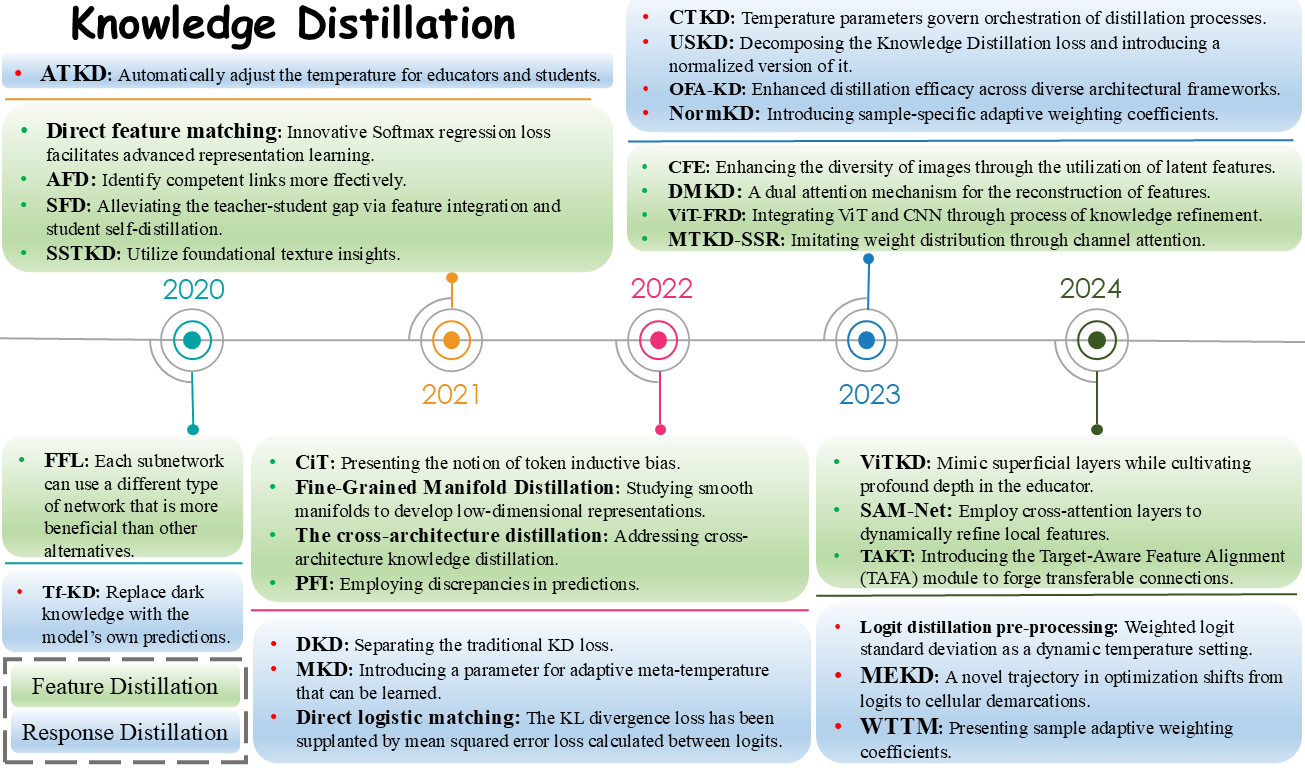}
    \caption{
    \textbf{Knowledge Distillation Exploration Timeline.}
    The figure collects and organizes the development of knowledge distillation methods in ViT model lightweighting. It includes two parts: feature knowledge distillation and response knowledge distillation.
 }
    \label{fig:vis_token_KD}
\end{figure*}

\section{\textbf{Knowledge Distillation}}
\label{Knowledge Distillation}
Knowledge distillation has emerged as a powerful approach to compress complex ViT models into lightweight and efficient variants   suitable for image recognition tasks.
By extracting and transferring knowledge from a large, pre-trained teacher model to a smaller student model, knowledge distillation enables the creation of compact ViT architectures that maintain high accuracy while significantly reducing storage requirements and inference time. 
\citep{hinton2015distillingknowledgeneuralnetwork,Gou2021-ty} propose the concept of knowledge distillation, mainly focused on extracting class probability distributions through softening of labels. 
In the context of lightweight online ViT strategies, knowledge distillation can be broadly categorized into two main topics: \textbf{Feature Knowledge Distillation} in Section~\ref{Feature Knowledge Distillation} and \textbf{Response Knowledge Distillation} in Section~\ref{Response Knowledge Distillation}. 
In Figure~\ref{fig:vis_token_KD}, we summarize the timeline of significant explorations on these two topics.
It should be noted that the relation distillation, present in CNN models, remains applicable, while the relation distillation among tokens in ViT is more adapted.

\subsection{\textbf{Feature Knowledge Distillation}}
\label{Feature Knowledge Distillation}
Feature-based knowledge distillation (FKD) focuses on transferring rich intermediate-level information from a teacher model to a student model, capitalizing on the detailed representations within the internal layers of teacher networks. 
FKD employs the multi-faceted strategy: \textbf{Attention Feature Distillation}, captures essential attention patterns, \textbf{Structured Feature Distillation}, compresses structural representations, and \textbf{Direct Feature Mimicking}, ensures that important features are transferred directly.

FKD emphasizes comprehensive guidance through intermediate feature information. 
The typical loss function for feature-based distillation is defined as:
\begin{equation}
    \begin{aligned}
        \mathcal{L}_{\text{FKD}}(F^{S}, F^{T}) = \mathcal{L}_{\text{D}}(\phi^{S}(F^{S}), \phi^{T}(F^{T})),
    \end{aligned}
\end{equation}
where \( F^{S} \) and \( F^{T} \) denote the intermediate feature mappings from the student and teacher, respectively. The functions \( \phi^S \) and \( \phi^T \) transform these features to include enhanced information such as attention mechanisms, activation boundaries, neuron selectivity, and probability distributions. The loss function \(\mathcal{L}_{\text{D}}\) measures the similarity between features, using metrics such as mean square error and Kullback-Leibler divergence.

\subsubsection{\textbf{Attention Feature Distillation}}
\label{Attention Feature Distillation}
Attention feature distillation transfers valuable information from teacher model attention maps.
AttnDistill \citep{wang2022attentiondistillationselfsupervisedvision} aligns class labels using the Projector Alignment (PA) module and approximates teacher network attention maps with the Attention Guidance (AG) module.
PFI \citep{li2022knowledge} guides feature mimicry by emphasizing high-variance areas based on prediction variance.
ViT-FRD \citep{fan2023vit} refines knowledge transfer by integrating ViT and CNN using the Collaborative Reformer based on Clustering (CCReformer), the Linear Transformer (Linformer) as the knowledge distillation modules.
CFD \citep{sun2023accelerating} employs a dataset-independent classifier to distill the distribution of refined features to the student.
AFD\footnote{\url{https://github.com/holger24/AFD}} \citep{ji2021show} takes advantage of attention-based meta-networks to control the strength of distillation by identifying similarities.
SAM-Net\footnote{\url{https://github.com/benjaminkelenyi/SAM-Net}} \citep{kelenyi2024sam} dynamically adapts local features based on environmental and spatial correlations by exploiting cross-attention.
Attention Probe-Based Distillation \citep{wang2022attentiondistillationselfsupervisedvision} selects significant real-world data and uses a probe knowledge distillation algorithm to train lightweight student models.

\begin{figure}[tb]
    \centering
    \includegraphics[width=0.65\linewidth]{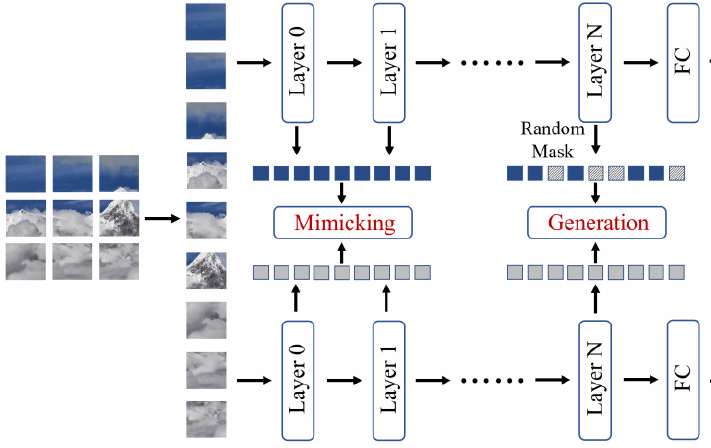}
    \caption{
    \textbf{Feature Knowledge Distillation Example}.
    ViTKD~\citep{yang2024vitkd} observes that regardless of model capacity, the attention gap in shallow layers is relatively small, advocating for a direct alignment strategy. In contrast, the attention distribution in deeper layers is more diverse, thus necessitating a generative alignment approach.
    }
    \label{fig:feature kd}
\end{figure}

\subsubsection{\textbf{Structured Feature Distillation}}
\label{Structured Feature Distillation}
Structured feature distillation ensures effective knowledge transfer while   maintaining structural integrity, which is crucial for lightweight vision transformers. Preserves and transfers structural information, helping student networks capture complex patterns and relationships, thus improving performance and generalization.

KDFAS~\citep{zhang2023kdfas} addresses the issue of embedding dimension gap by employing a covariance matrix during feature-level knowledge transfer, while DMKD~\citep{yang2024dmkd} employs a dual-attention mechanism to guide spatial and channel information to its   respective masking branches and integrates reconstructed features. 
PromptKD~\citep{li2024promptkd} transfers category embedding from teacher to student through shared distillation. 
CrossKD~\citep{wang2024crosskd} makes the student generate cross head predictions like the teacher.
MTKD-SSR\footnote{\url{https://github.com/ChaofWang/Awesome-Super-Resolution}}~\citep{gou2023multi} introduces the Staged Channel Distillation (SCD) mechanism, which takes advantage of channel attention to allow the student network to mimic the weight distribution within    the channel feature map. 
PSD\footnote{\url{https://github.com/webtoon/psd}}~\citep{zhu2023learn} incorporates multiple self-distillations with a shared embedded network between the teacher and the student during each process, and TAKT~\citep{Xiong2023TAKTTK} proposes the Target-Aware Feature Alignment (TAFA) module to establish transferable feature connections between the source and target domains. 
The cross-architecture knowledge distillation approach~\citep{liu2022cross} aligns intermediate features using projectors for partial cross-attention and grouped linear   transformations, while Fine-Grained Manifold Distillation~\citep{ni2022manifold} creates low-dimensional features by learning smooth manifolds embedded within the original feature space. 
DearKD~\citep{chen2022dearkd} is a two-stage framework that initially refines the inductive bias from the early intermediate layers of the CNN, subsequently allowing the transformer to be fully leveraged through unrefined training.

\subsubsection{\textbf{Direct Feature Mimicking}}
\label{Direct Feature Mimicking}
Diret Feature Mimicking alleviates the need for complex preprocessing of alignment targets, helping to preserve the native quality and inherent patterns of the information~\citep{Gao2024-mv}.
As shown in Figure~\ref{fig:feature kd}, ViTKD\footnote{\url{https://github.com/yzd-v/cls_KD}}~\citep{yang2024vitkd} employs ``linear layers" and ``correlation matrices" for shallow tasks, a convolutional projector for deeper \quad generative tasks. MMKD~\citep{zhang2023adaptive} uses a meta-weighting network to integrate logic and intermediate features from various teacher models to guide the student network. \citep{lin2022knowledge} proposes a one-to-one spatial matching method that employs a target-aware transformer\footnote{\url{https://github.com/Kevoen/TATrack}} (TaT) to align the feature component.

\subsubsection{\textbf{Summary: Feature Knowledge Distillation}}
To improve the performance of small capacity ViT models and facilitate their deployment in constrained environments, directly training compact ViT models often fails to achieve satisfactory performance, and simple knowledge distillation methods are insufficient to fully transfer the rich information embedded within the teacher model. 

\begin{figure}[tb]
    \centering
    \includegraphics[width=0.65\linewidth]{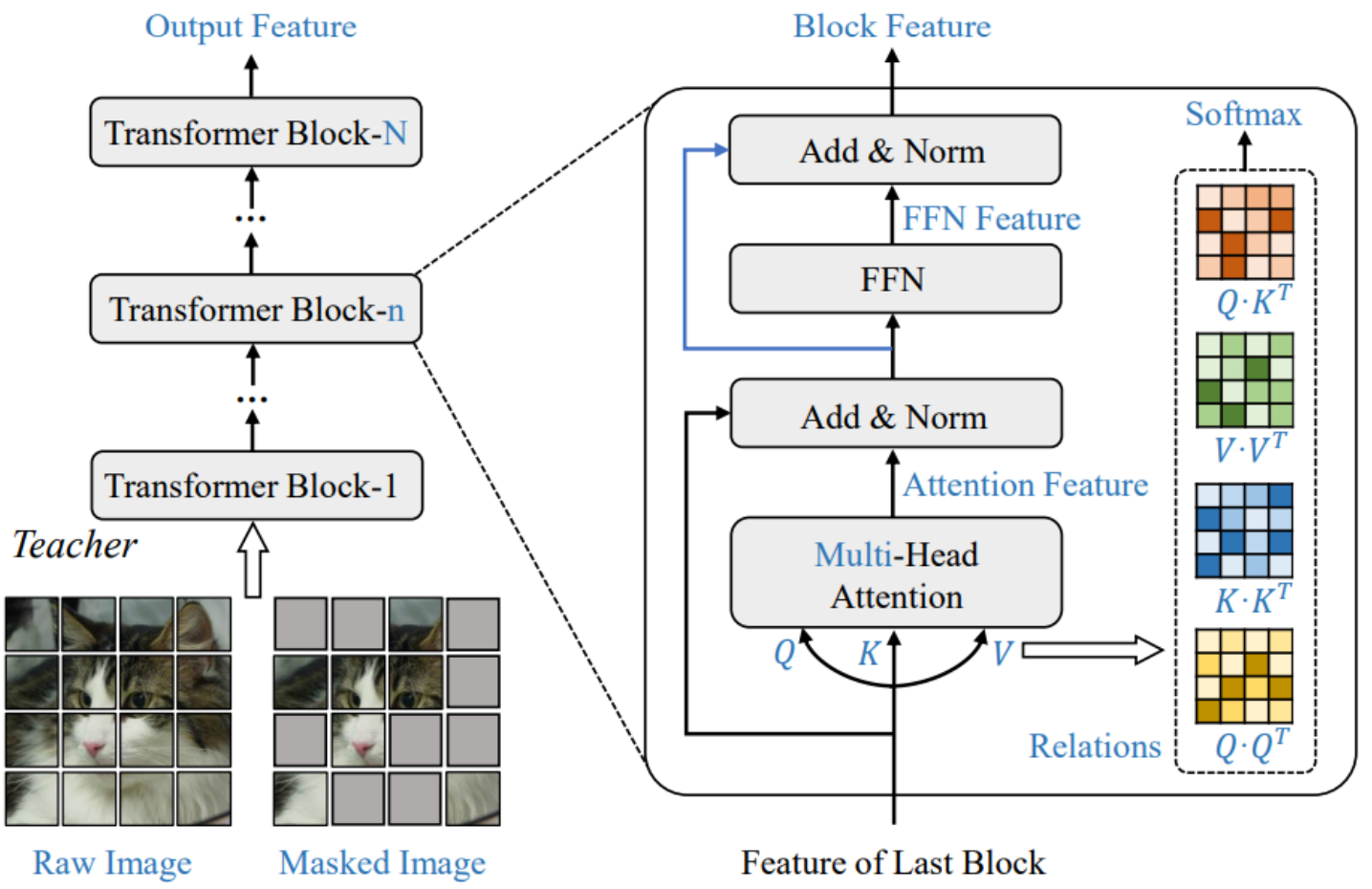}
    \caption{
    \textbf{Token Relation Transfer.}
    TinyMIM~\citep{ren2023tinymim} extracts the relation knowledge among tokens, including Q-K, V-V, K-K, and Q-Q, which differs from the conventional CNN of capturing the relation knowledge among images within the batch.
    }
    \label{fig:token re}
\end{figure}

Therefore, extracting the abundant latent patterns from the teacher network presents a promising direction.
Attention Feature Distillation enhances the ability of lightweight ViT models to identify critical information, while Structured Feature Distillation maintains the integrity of knowledge structures during the distillation process. Direct Feature Mimicking alleviates the burden of complex preprocessing.
This multifaceted feature knowledge distillation strategy effectively transfers the rich intermediate-layer information. 
As illustrated in Figure~\ref{fig:token re}, TinyMIM\footnote{\url{https://github.com/OliverRensu/TinyMIM}}~\citep{ren2023tinymim}  takes advantage of the unique representational advantages inherent in ViT architectures and
extracts relation knowledge among tokens as a complementary feature knowledge transfer, which differs from the conventional CNN approach of extracting relation knowledge among images within the input batch.

\subsection{\textbf{Response Knowledge Distillation}}
\label{Response Knowledge Distillation}
Feature-based approaches recently come to the fore due to its superior performance in numerous applications. Consequently, response knowledge distillation (RKD) is relatively neglected. RKD, which facilitates direct learning from the output layer of the teacher model by the student model, presents distinct advantages, including rapid learning and straightforward implementation, and the potential remains largely unexplored. This outcome-driven approach makes it widely applicable across various tasks, given the recognition task, the soft probability distribution $p$ is defined as:
\begin{equation}
p(z_i ; T) = \frac{\exp(z_i / T)}{\sum_{j=1}^{N} \exp(z_j / T)},
\end{equation}
where \(z_i\) is the logit for class \(i\), \(N\) is the number of classes, and \(T\) is a temperature parameter controlling the distribution smoothness.  
RKD aims to align the teacher probability distribution, \(p(z^T; T)\), with the student, \(p(z^S; T)\), by minimizing a distance function \(\mathcal{L}_{dis}\):
\begin{equation}
\mathcal{L}_{RKD} = \mathcal{L}_{dis}\left(p(z^S; T), p(z^T; T)\right),y
\end{equation}
where \(\mathcal{L}_{dis}\) can be instantiated with various distance metrics, such as Kullback-Leibler divergence,    mean squared error, or Pearson correlation. For example, the classic KD loss is formulated as:
\begin{equation}
\mathcal{L}_{RKD} =  \sum_{i=1}^{N} p(z_i^{T} ; T) \log \frac{p(z_i^{T} ; T)}{p(z_i^{S} ; T)},
\end{equation}
where \(z_i^T\) and \(z_i^S\) represent the teacher and student logits for class \(i\), respectively. 

Improving response-based knowledge distillation can involve adjusting distillation \textbf{Temperature Variants}  to transfer information-rich knowledge, optimizing \textbf{ Objective Function Design}  to diversify learning objectives, and \textbf{Empowering Soft Labels}  to prevent overfitting. 

\subsubsection{\textbf{Temperature Variants}}
\label{Temperature Variants}
Temperature variation by modulating the softness of the teacher model logit output, temperature scaling provides smoother and more informative probability distributions to the student model. 
However, traditional fixed-temperature methods often fail to exploit the diverse properties of logit distributions across samples, leading to suboptimal knowledge transfer~\citep{Zhao2024-wr}.

To overcome these limitations, there are several advanced techniques. 
ATKD~\citep{guo2022reducing} dynamically adjusts the temperatures for both teacher and student outputs, balancing clarity and mitigating performance degradation. 
MKD~\citep{liu2022metaknowledgedistillation} introduces meta-learned temperature parameters that adapt throughout the training process based on the learning objective gradient. 
\citep{jin2023multi} introduces a multi-level logit distillation framework, which transforms them into multiple outputs with varying temperatures, and   performs multi-level alignment  matching. 
NormKD~\citep{chi2023normkdnormalizedlogitsknowledge} customizes the temperatures for each sample to achieve a uniform distribution by analyzing logit output.
SKD~\citep{yuan2024student} integrates a temperature-scaled LogSoftmax function with a learning simplifyer to improve distillation loss during joint training. 
\citep{sun2024logit} propose using a weighted logit standard deviation as the adaptive temperature, combined with Z-score preprocessing to normalize logits before applying Softmax.


\subsubsection{\textbf{Objective Function Design}}
\label{Objective Function Design}
Creative formulations of objective functions enhance the distillation process, allowing a wider variety of training for compact student models.

Decoupled Knowledge Distillation (DKD)~\citep{zhao2022decoupled} splits KD loss into Target Class (TCKD) and Non-Target Class (NCKD) components, with NCKD weighting independent of teacher confidence. 
Unified Self-Knowledge Distillation (USKD)~\citep{yang2023knowledge} further refines the KD loss into target and non-target losses using cross-entropy, normalizes non-target logits to formulate the non-target knowledge distillation (NKD) loss, and generates customized soft labels to potentially improve knowledge transfer.
PTLoss~\citep{zhang2023not} represents the original KL-based distillation loss function through a maclaurin series and implicitly transforms the original teacher into a proxy teacher by perturbing the first-order term in this series.
NTCE~\citep{li2024ntce} introduces a Magnitude Kullback Leibler (MKL) divergence to better represent non-target categories, combined with diversity-based data augmentation (DDA) to enrich non-target category diversity.
CSKD~\citep{zhao2023cumulative} terminates the pooling operation following the final feature map, producing dense predictions for each corresponding feature location. From these results, hard labels are derived to serve as target labels, utilizing cross-entropy as the loss function for spatial knowledge transfer.


\subsubsection{\textbf{Empowering Soft Labels}}
\label{Empowering Soft Labels}
Reducing the influence of the teacher model in knowledge distillation seeks to create vision transformer models that are more lightweight and autonomous. This approach moves away from a strong reliance on pre-trained teachers, allowing students to learn on their own.

The Tf-KD framework~\citep{yuan2020revisiting} replaces dark knowledge with the model own predictions, using self generated soft targets for regularization, and achieves smoothing effects similar to label smoothing regularization (LSR). 
MTKDSSR~\citep{gou2023multi} incorporates staged channel distillation, staged response distillation, and cross-stage distillation, using student self-reflection  through self-distillation of response-based knowledge. 
The OFA-KD framework~\citep{hao2024one} eliminates architecture-specific information and employs exit branches in the student model trained with teacher model logits.
MEKD~\citep{ma2024aligning} transitions from logits to cell boundaries, innovating with de-privatization and distillation    for lightweight model transformations. 
WTTM~\citep{zheng2024knowledgedistillationbasedtransformed} enhances the training of the student model using sample-adaptive weighting to emulate the power-transformed probability distributions of the teacher. Tiny-ViT~\citep{wu2022tinyvit} focuses on reducing memory and computational overhead through logit sparsification and efficient knowledge transfer during pre-training.

\subsubsection{\textbf{Summary:Response Knowledge Distillation}}
Response knowledge distillation enhances efficiency, prevents overfitting, and facilitates cross-architecture knowledge sharing due to its flexible implementation~\citep{Liu2024-fp}.
We cover three topics: Temperature Variants exploit logit distribution characteristics better than fixed temperature methods; Objective Function Design diversifies learning objectives for more comprehensive knowledge transfer; Empowering Soft Labels lessens dependence on pre-trained teacher models, enhancing student models independent learning.
Furthermore,
\textit{\textbf{(i).}}  Response knowledge distillation provides soft guidance, offering richer information than hard labels. 
\textit{\textbf{(ii).}}  Given the prevalent long-tail issue in image recognition tasks, leveraging the teacher model performance on rare categories to improve the student model's learning can be achieved by adjusting the response distillation loss weight to focus on rare categories. 
\textit{\textbf{(iii).}} Provides a lightweight response-based multi-scale knowledge transfer pathway~\citep{wei2024scaled}, distinctly different from the burdensome multi-scale feature knowledge distillation.

\subsection{\textbf{Knowledge Distillation Discussion}}
Knowledge distillation bridges CNN and ViT models for cross-structural knowledge transfer. We focus on KD strategies that take advantage of both, especially ViT. For example, AttnDistill~\citep{{wang2022attentiondistillationselfsupervisedvision}} uses the importance distribution between the [CLS] token and the image tokens to achieve a more efficient alignment of the target. ViT models effectively extract relevant knowledge for tasks in response knowledge distillation. 
TinyMIM~\citep{ren2023tinymim} adapts the popular relation knowledge transfer in CNN to ViT using Q-K, V-V, K-K, and Q-Q relationships among tokens.
We summarize the limitations of existing methods as follows:
\begin{enumerate}[label=\alph*.]
    \item Exploiting the unique structural advantages of ViT for knowledge distillation, including head-level distillation, where existing methods still involve numerous artificial settings or task-coupling characteristics.
    \item The fast distillation direction, focusing on response knowledge distillation, can be extended to feature knowledge distillation and even combined with self-supervised optimization objectives.
    \item  Addressing privacy and security concerns during the knowledge distillation process as well as enhancing the robustness of the student network in abnormal scenarios post-distillation.
\end{enumerate}

\section{\textbf{Future Research Directions}}
\label{Future Research Directions}
Vision transformer models excel in computer vision, but their high parameter and computational needs hinder their practical use. 
Future lightweight ViT models will emphasize efficiency, intelligence, and security.

\noindent 
\textbf{{Representation Advantage Expansion.}} Breaking free of the limitations of static model structures, the focus shifts from the previous global perspective to a more fine-grained local perspective, while integrating the unique representational advantages of ViT, such as deep adaptability and response knowledge distillation, which remain directions worthy of profound exploration in their ingenious integration with ViT.

\noindent 
\textbf{{End-to-End Training System.}} 
Offline lightweight ViT techniques, such as post-training quantization, remain relatively independent of online lightweight techniques. Future exploration can focus on designing end-to-end training systems, where errors arising from token pruning and merging can also be perceived and controlled.

\noindent 
\textbf{{Automatic Configuration.}} 
Existing online lightweight strategies for ViT still involve a significant number of scenarios that require manual intervention. We advocate for the automation of these processes, focusing primarily on two aspects:
\begin{enumerate}[label=\alph*.]
    \item Analysis of performance bottlenecks and selection of the corresponding combination strategies can be automated by structuring the pipeline.
    \item Within specific strategies, manually set parameters, such as the location of the layer and the corresponding rate, remain for token pruning and merging.
\end{enumerate}

\noindent 
\textbf{{Privacy-Sensitive Lightweight.}} 
Traditional knowledge distillation requires access to the original training dataset, which may not be feasible in privacy-sensitive application scenarios. Data-Free Knowledge Distillation (DFKD) methods generate synthetic data using only the teacher model, without accessing the original data, thus protecting data privacy. However, DFKD has not yet been explored in the ViT domain. We offer several valuable research directions:
\begin{enumerate}[label=\alph*.]
    \item Using the diffusion vision transformer~\citep{rombach2022high,xiang2024dkdmdatafreeknowledgedistillation} instead of the CNN generator to transfer knowledge on synthetic datasets that better match real data distributions.
    \item The current DFKD setup~\citep{tran2024nayer} still reveals the original dataset input resolution to the user, and using the flexible ViT input form can enhance the benefits of the data-free paradigm.
\end{enumerate}

\noindent 
\textbf{{Inference Security.}} 
Well-trained teacher networks possess greater robustness against anomalous inputs, but the lightweight optimization process of the ViT model does not explicitly transfer this knowledge. This makes small ViT models more susceptible to adversarial attacks. Consequently, the extraction and transfer of robust, interference-resistant knowledge is a crucial consideration for future practical deployments.

\section{\textbf{Conclusion}}
In this paper, we conduct a comprehensive investigation into lightweight techniques for vision transformer in image recognition,  with particular emphasis on online lightweight strategies. 
Our exploration is divided into three key topics: vision transformer \textbf{Efficient Component Design}, \textbf{Dynamic Network}, and \textbf{Knowledge Distillation}. 
We analyze these lightweight approaches both quantitatively and qualitatively in terms of the trade-offs between precision and efficiency, under the consistent view on ImageNet-1K. 
For each topic, we present the methods in the form of input flow forward the network, allowing flexible component combinations and bottleneck identification.
In addition, we discuss the limitations of related explorations within each topic and propose potential solutions. However, numerous unknown challenges remain in the face of complex real-world scenarios, thus advocate for maintaining the privacy and security of the lightweighting process to ensure the broad acceptance and usage of ViT. 
In summary, we hope that this survey will serve as a valuable resource for researchers, providing insight and guidance for the development of lightweight vision systems.

\bibliographystyle{ACM-Reference-Format}
\bibliography{main}

\end{document}